\RequirePackage{etex} 
\documentclass{article}

\usepackage[preprint,nonatbib]{neurips_2021}

\usepackage[utf8]{inputenc} % allow utf-8 input
\usepackage[T1]{fontenc}    % use 8-bit T1 fonts
\usepackage{hyperref}       % hyperlinks
\usepackage{url}            % simple URL typesetting
\usepackage{booktabs}       % professional-quality tables
\usepackage{amsfonts}       % blackboard math symbols
\usepackage{nicefrac}       % compact symbols for 1/2, etc.
\usepackage{microtype}      % microtypography

\usepackage{amsmath}
\usepackage[cmintegrals]{newtxmath}
\usepackage{booktabs}       % professional-quality tables
\usepackage{threeparttable}
\usepackage[lambda,advantage,operators,sets,adversary,landau,probability,notions,logic,ff,mm,primitives,events,complexity,asymptotics,keys]{cryptocode}

\newtheorem{definition}{Definition}

\usepackage[titles]{tocloft}

\title{Privacy-Preserving Machine Learning: Methods, Challenges and Directions}

\author{%
  Runhua Xu$^{1}$\thanks{Part of this work was done while Runhua Xu was at the School of Computing and Information, University of Pittsburgh.},  Nathalie Baracaldo$^{1}$, and James Joshi$^{2}$\thanks{This work was performed while James Joshi was serving as a program director at NSF; and the work represents the views of the authors and not that of NSF.} \\
  \\
  $^{1}$ IBM Research - Almaden Research Center, San Jose, CA, United States, 95120\\
  $^{2}$ School of Computing and Information, University of Pittsburgh,   Pittsburgh, PA, United States, 15260 \\
  \texttt{runhua@ibm.com, baracald@us.ibm.com, jjoshi@pitt.edu} \\
}

\begin{document}

\maketitle

\begin{abstract}
Machine learning (ML) is increasingly being adopted in a wide variety of application domains. Usually, a well-performing ML model
% , especially, emerging deep neural network model, 
relies on a large volume of training data and high-powered computational resources. Such a need for and the use of  huge volumes of data raise serious privacy concerns because of the potential risks of leakage of highly privacy-sensitive information; further, the evolving regulatory environments that increasingly restrict access to and use of privacy-sensitive data add significant challenges to fully benefiting from the power of ML for data-driven applications.
A trained ML model may also be vulnerable to adversarial attacks such as membership, attribute, or property inference attacks and model inversion attacks.  
Hence, well-designed privacy-preserving ML (PPML) solutions are critically needed for many emerging applications.
Increasingly, significant research efforts from both academia and industry can be seen in  PPML areas that aim toward integrating privacy-preserving techniques into ML pipeline or specific algorithms, or designing various PPML architectures.
In particular, existing PPML research cross-cut ML, systems and applications design, as well as security and privacy areas; hence, there is a critical need to understand state-of-the-art research, related challenges and a research roadmap for future research in PPML area.
In this paper, we systematically review and summarize existing privacy-preserving approaches and propose a \textit{\textbf{P}hase}, \textit{\textbf{G}uarantee}, and \textit{\textbf{U}tility} (\textit{\textbf{PGU}}) triad based model to understand and guide the evaluation of various PPML solutions by  decomposing their privacy-preserving functionalities.
We discuss the unique characteristics and challenges of PPML and outline possible research directions that leverage as well as benefit multiple research communities such as ML, distributed systems, security and privacy.  
\\ 
\\
\textit{\textbf{Key Phrases:}} Machine Learning, Privacy-Preserving Machine Learning
\end{abstract}

% \begin{IEEEkeywords}
% survey, privacy-preserving methodologies, machine learning
% \end{IEEEkeywords}

\newpage
\tableofcontents
\newpage

\section{Introduction}

% ML background and need for data and computation
Machine learning (ML) is increasingly being applied in a wide variety of application domains. 
For instance, emerging deep neural networks, also known as deep learning (DL), have shown significant improvements in model accuracy and performance, especially in application areas such as computer vision, natural language processing, and speech or audio recognition \cite{lecun2015deep, goodfellow2016deep, rahwan2019machine}.
Emerging federated learning (FL) is another collaborative ML technique that enables training a high-quality model while training data remains distributed over multiple decentralized devices \cite{mcmahan2016communication, konevcny2016federated}. 
FL has shown its promise in various application domains, including healthcare, vehicular networks, intelligent manufacturing, among others \cite{xu2019federated,samarakoon2019distributed,hao2019efficient}. 
Although these models have shown considerable success in AI-powered or ML-driven applications, they still face several challenges, such as (\romannumeral1) lack of powerful computational resources and (\romannumeral2) availability of huge volumes of data for model training.
In general, the performance of an ML system relies on a large volume of training data and high-powered computational resources to support both the training and inference phases.

To address the need for computing resources with high-performance CPUs and GPUs, large memory storage, etc., existing commercial ML-related infrastructure service providers, such as Amazon, Microsoft, Google, and IBM, have devoted significant amounts of their efforts toward building \textit{infrastructure as a service} (IaaS) or \textit{machine learning as a service} (MLaaS) platforms with appropriate rental fees.
The resource-limited clients can employ ML-related IaaS or MLaaS to manage and train their models first and then provide data analytics and prediction services through their applications directly.

Availability of massive volumes of training data is another challenge for ML systems.
Intuitively, more training data indicates better performance of an ML model; thus, there is a need for collecting large volumes of data and in many cases from multiple sources.
However, the collection and use of the data, as well as the creation and use of ML models, raise serious privacy concerns because of the risks of leakage of private or confidential information. 
For instance, recent data breaches have significantly increased the privacy concerns of large-scale collection and use of personal data \cite{rosati2019social, vemprala2019social}. 
An adversary can also infer private information by exploiting an ML model via various inference attacks such as membership inference attacks \cite{shokri2017membership,bernau2019assessing,jia2019memguard,li2020membership,nasr2018machine}, model inversion attacks \cite{fredrikson2015model,he2019model,wu2016methodology}, property inference attacks \cite{ganju2018property,parisot2021property}, and privacy leakage from gradients exchanged in distributed ML scenarios \cite{zhu2019deep,zhao2020idlg}.
For instance, in an membership inference attack, an attacker can infer whether or not data related to a particular patient has been included in the training of an HIV-related ML model.
In addition, existing regulations such as the Health Insurance Portability and Accountability Act (HIPPA) and more recent ones such as the European General Data Protection Regulation (GDPR), Cybersecurity Law of China, California Consumer Privacy Act (CCPA), etc., increasingly restrict the availability and use of privacy-sensitive data.
Such privacy concerns and challenges pose significant roadblocks to the adoption of ML models for real-world applications.

To tackle the increasing privacy concerns related to using ML in applications, 
in which users' privacy-sensitive data such as electronic health/medical records, location information, etc., are stored and processed, 
it is crucial to devise innovative privacy-preserving ML (PPML) solutions.
More recently, there have been increasing efforts focused on PPML research that integrate existing anonymization mechanisms into ML pipelines or design innovative new privacy-preserving methods and architectures for ML systems.
Recent surveys focused on ML, including Federated Learning  such as in \cite{kairouz2019advances,pouyanfar2018survey,li2019federated,yang2019federated,lyu2020threats, yin2021comprehensive} partially illustrate or discuss the specific privacy and security issues in ML or FL systems.
Each existing PPML approach addresses part of privacy concerns or is only applicable to limited scenarios.
There is no unified or or holistic view of PPML solutions.
For instance, the adoption of differential privacy in ML systems can lead to model utility loss, e.g., reduced model accuracy.
Similarly, the use of secure multi-party computation approaches incurs high communication overhead or computation overhead. 
The communication overhead is caused by transmitting a large volume of intermediate data, such as garbled tables of circuit gates. 
At the same time, the adoption of advanced cryptosystems \cite{gentry2009fully, boneh2011functional} leads to computation overhead.
% Similarly, the use of secure multi-party computation approaches incurs high communication overhead because of a large volume of intermediate data such as garbled tables of circuit-gates that need to be transmitted during the execution of the protocols or high computation overhead due to the adopted ciphertext-computational cryptosystems.

\begin{figure*}[t]
    \centering
    \includegraphics[origin=c, width=\textwidth]{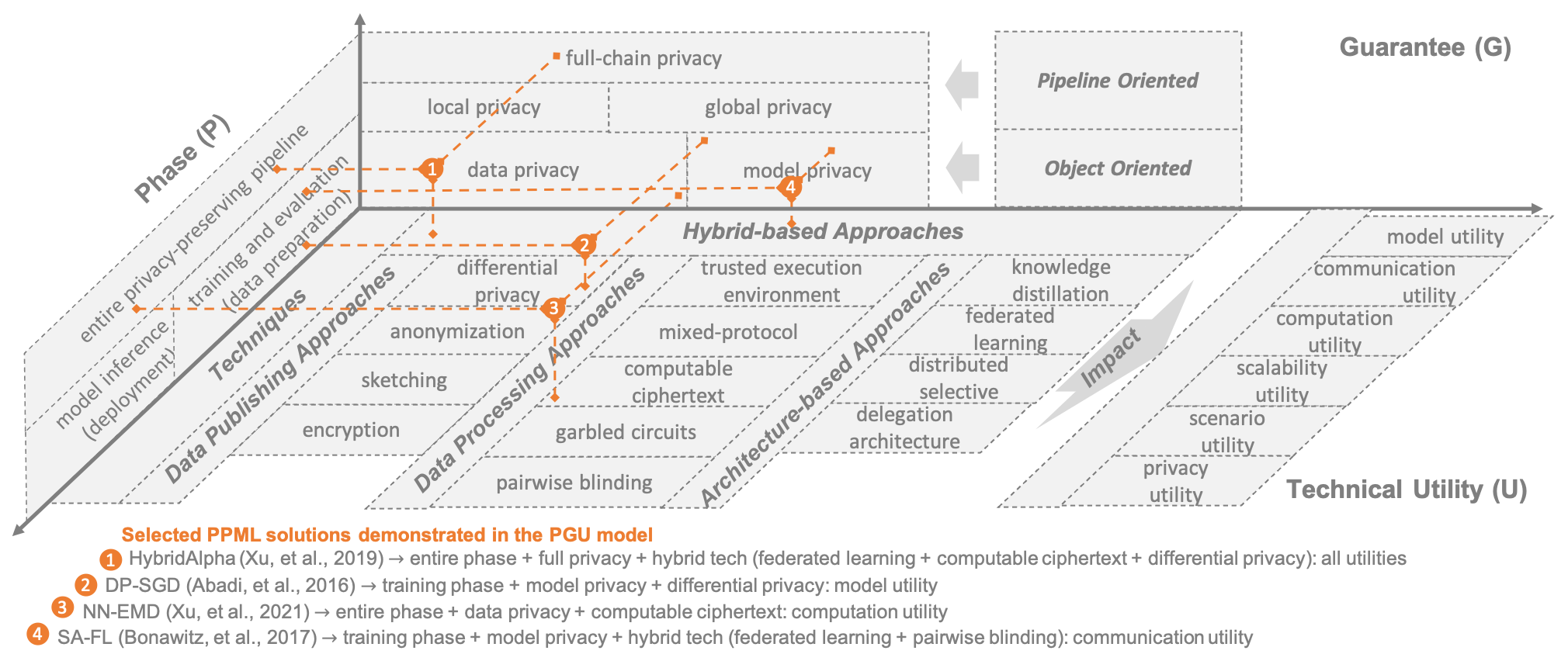}
    \caption{An overview of PGU model to evaluate the privacy-preserving machine learning systems and illustration of selected PPML examples in the PGU model. The demonstrated PPML examples in the figure are HybridAlpha \cite{xu2019hybridalpha}, DP-SGD \cite{abadi2016deep}, NN-EMD\cite{xu2021nn}, SA-FL\cite{bonawitz2017practical}.}
    \label{fig:pgu}
\end{figure*}

% As the systematization of knowledge discussion in \cite{papernot2018sok} and the survey of ML security issues and corresponding countermeasures in \cite{barreno2006can, barreno2010security}, the security issues such as stealing the ML models and injecting Trojans have been thoroughly discussed.
ML security, such as issues of stealing the ML models, injecting Trojans and availability of ML services and corresponding countermeasures, have been discussed in various recent articles such as those related to the systematization of knowledge \cite{papernot2018sok},  or surveys/analyses \cite{barreno2006can, barreno2010security,nasr2019comprehensive}.
However, there is still a lack of systematization of knowledge discussion and evaluation with privacy as the key focus.  
Inspired by the CIA (confidentiality, integrity and availability) triad in security  designed to more holistically understand information security,
in this paper, we propose a \textit{\textbf{PGU}} triad - referring to \textit{\underline{P}hase}, \textit{\underline{G}uarantee}, and \textit{technical \underline{U}tility} -  to better comprehend and help guide the  understanding or evaluation of various PPML solution space by decomposing functionalities/features of privacy-preserving approaches, as illustrated in \figurename~\ref{fig:pgu}.
Here, \textit{phase} represents the phase of privacy-preserving functionalities that occur in various phases of a ML pipeline;
\textit{guarantee} denotes the  strength or scope of the privacy protection under a given set of threat models and/or trust assumptions;
\textit{utility} captures the impact of adopted privacy solutions  on the accuracy or usefulness of computational results of ML systems.
Based on the PGU analysis framework, we also discusses various challenges and potential future research directions in PPML.

More specifically, we first introduce the general ML pipeline in a nutshell. Then we discuss the PPML pipeline from various phases of privacy-preserving functionalities that occur in the process-chain in these systems; these include \textit{privacy-preserving data preparation}, \textit{privacy-preserving model training and evaluation}, \textit{privacy-preserving model deployment}, and \textit{privacy-preserving model inference}.

Following that, we discuss the privacy guarantees provided by existing PPML solutions by analyzing the strength and/or scope of the privacy protection from two perspectives: object-oriented privacy protection and pipeline-oriented privacy protection, based on common threat model settings and trust assumptions.
From an object-oriented perspective, PPML solutions either aim to protect \textit{input privacy} by preventing the leakage/exposure of private information from training or inference data samples, or to protect \textit{model privacy} by mitigating privacy disclosure from the learned model.
From a pipeline viewpoint, PPML solutions are concerned with the privacy-preserving functionality associated with an entity or collection of entities in the pipeline of a machine learning solution.
% Usually, it includes \textit{local privacy}, \textit{global privacy}, and \textit{full-chain privacy}, which are discussed later.
Typically, it encompasses local privacy, global privacy, and full-chain privacy, all of which will be described in further detail later.

% In addition, we evaluate various PPML systems by investigating their technical utility.
Additionally, we investigate the technical utility of various PPML systems.
% In particular, we first discuss the underlying privacy-preserving approaches by decomposing and classifying existing PPML solutions into four categories: 
We begin by dissecting and classifying existing PPML solutions into four categories: 
\textit{data publishing} approaches, \textit{data processing} approaches, \textit{architecture based} approaches and \textit{hybrid} approaches.
Then, we examine their impact on an ML system's utility, including \textit{computation utility}, \textit{communication utility}, \textit{model utility}, \textit{scalability utility}, \textit{scenario utility}, among others.

Finally, we discuss the challenges of designing PPML solutions and future research directions.

\noindent\textbf{Organization}.
The remainder of this paper is organized as follows.
We briefly present the ML pipeline in Section~\ref{sec:overview} by reviewing the critical tasks in ML-related systems, and third-party facility-related ML solutions.
In Section~\ref{sec:phase}, we present general discussion of existing privacy-preserving methods by considering the phases where they are applied, and discuss privacy guarantees in Section~\ref{sec:pg}.
We investigate technical utility of PPML solutions in Section~\ref{sec:tu} by summarizing and classifying privacy-preserving techniques and their impact on ML systems.
Furthermore, we also discuss the challenges and open problems in PPML solutions and outline the promising directions of future research in Section~\ref{sec:cr}.
Finally, we conclude the paper   in Section~\ref{sec:conclusion}.

\section{Machine Learning Pipeline in a Nutshell}
\label{sec:overview}

The four phases of a machine learning system are typically as follows: 
\textit{data preparation or preprocessing}, \textit{model training and evaluation}, \textit{model deployment}, and \textit{model inference}. 
More broadly, the ML pipeline can be divided into \textit{training} and \textit{serving} phases. 
% In such a case, \textit{training} encompasses processes such as collecting data and preprocessing it, training a model, and evaluating the model performance; and  \textit{model serving} mainly focuses on using a trained model, such as how to deploy the model and provide inference services. 
In this scenario, model \textit{training} encompasses processes such as data collection and preprocessing, model training, and model evaluation; while model \textit{serving} mostly focuses on how to use a trained model, such as how to deploy the model and infer the result given a certain data sample.

In this section, we first formally differentiate computational tasks between the model training and model serving, and then based on this, we discuss privacy-preserving training and privacy-preserving serving approaches. 
Following that, we demonstrate a general machine learning pipeline that utilizes both self-owned and third-party infrastructure to handle the majority of machine learning-based workloads.
The processing pipeline is divided into three trust domains: \textit{trusted data owner}, \textit{trusted third-party}, and \textit{trusted model user}.
On this basis, we may examine the privacy guarantees provided by existing PPML solutions by considering various trust assumptions and probable adversarial threats.

\subsection{Computation Tasks in Model Training and Serving}
\label{sec:overview:cmp}

From the perspective of underlying computation tasks, there is no strict boundary between the model training and the model serving (i.e., inference) phases.
The computed function in the serving procedure could be viewed as a simplified version of the training procedure without loss computation, regularization, (partial) derivatives, and model weights update, which are needed during training phase.
For instance, in a stochastic gradient descent (SGD) based training approach, the computation that occurs at the inference phase could be viewed as one round of computation in the training phase without operations related to model gradients update.
In a more complex neural networks, the computation involved during the training phase includes continuously feeding  a set of data to the designed network for multiple training epochs. In contrast, the inference service can be treated as only one epoch of computation for one data sample to predict a label without a propagation procedure and related a regularization or normalization step.

Formally, given a set of training samples denoted as $(\pmb{x}_1, y_1),$ $...,$ $(\pmb{x}_n, y_n)$, where $\pmb{x}_i \in \mathbb{R}^{m}, y_i \in \mathbb{R}$, the goal of a ML model training (for simplicity, assume a linear model) is to learn a fit function denoted as 
\begin{equation}
    f_{\pmb{w},b}(\pmb{x}) = \pmb{w}^{\intercal}\pmb{x} + b,
\end{equation}
where $\pmb{w} \in \mathbb{R}^{m}$ is the set of model parameters, and $b$ is the intercept.
To find proper model parameters, usually, we need to minimize the regularized training error given by
\begin{equation}
    \label{eq:loss}
    E(\pmb{w}, b) = \frac{1}{n}\sum^{n}_{i=1}\mathcal{L}(y_i, f(\pmb{x}_i)) + \alpha R(\pmb{w}),
\end{equation}
where $\mathcal{L}(\cdot)$ is a loss function that measures model fit and $R(\cdot)$ is a regularization term (a.k.a., penalty) that penalizes model complexity; $\alpha$ is a non-negative hyperparameter that controls the regularization strength.
Regardless of various choices of $\mathcal{L}(\cdot)$ and $R(\cdot)$, stochastic gradient descent (SGD) is a common optimization method for unconstrained optimization problems.
A simple SGD method \textit{iterates} over the training samples and for each sample updates the model parameters according to the update rule given by
\begin{align}
    \pmb{w} &\gets \pmb{w} - \eta\nabla_{\pmb{w}} E = \pmb{w} - \eta[\alpha\nabla_{\pmb{w}} R + \nabla_{\pmb{w}}\mathcal{L}]\\
    b &\gets b - \eta\nabla_{b} E = b - \eta[\alpha\nabla_{b} R + \nabla_{b}\mathcal{L}]
\end{align}
where $\eta$ is the learning rate which controls the step-size in the parameter space. 

Given the trained model $(\pmb{w}_{\text{trained}},b_{\text{trained}})$, the goal of the model serving is to predict a value $\hat{y}$ for target sample $\pmb{x}$ as follows:
\begin{equation}
\label{eq:predict}
\hat{y} = f_{\pmb{w}_{\text{trained}}, b_{\text{trained}}}(\pmb{x}).
\end{equation}

As illustrated above, the computed functions in the inference phase (i.e., Equation \eqref{eq:predict}), is part of computed procedures in the SGD training (i.e., Equation \eqref{eq:loss}).
Similarly, in the case of a deep neural network,
the model training process is a model inference process with extra back-propagation to compute the partial derivative of weights. 
This also indicates that the task of privacy-preserving training is more challenging than the task of privacy-preserving serving.
% Most existing computation-oriented privacy-preserving training solutions indicate the availability to achieve privacy-preserving serving even though the proposal does not evidently state that, but not vice versa.
While the majority of extant privacy-preserving training solutions that rely on secure computation approaches imply the possibility of achieving privacy-preserving serving even when the proposal does not explicitly declare it; this is not true for vice versa.
A more specific demonstration is presented in Section~\ref{sec:tu:type2}.
% in the rest of the paper.

\subsection{An Illustration of Trusted Third-party based ML Pipeline}

\begin{figure*}[t]
    \centering
    \includegraphics[origin=c,width=\textwidth]{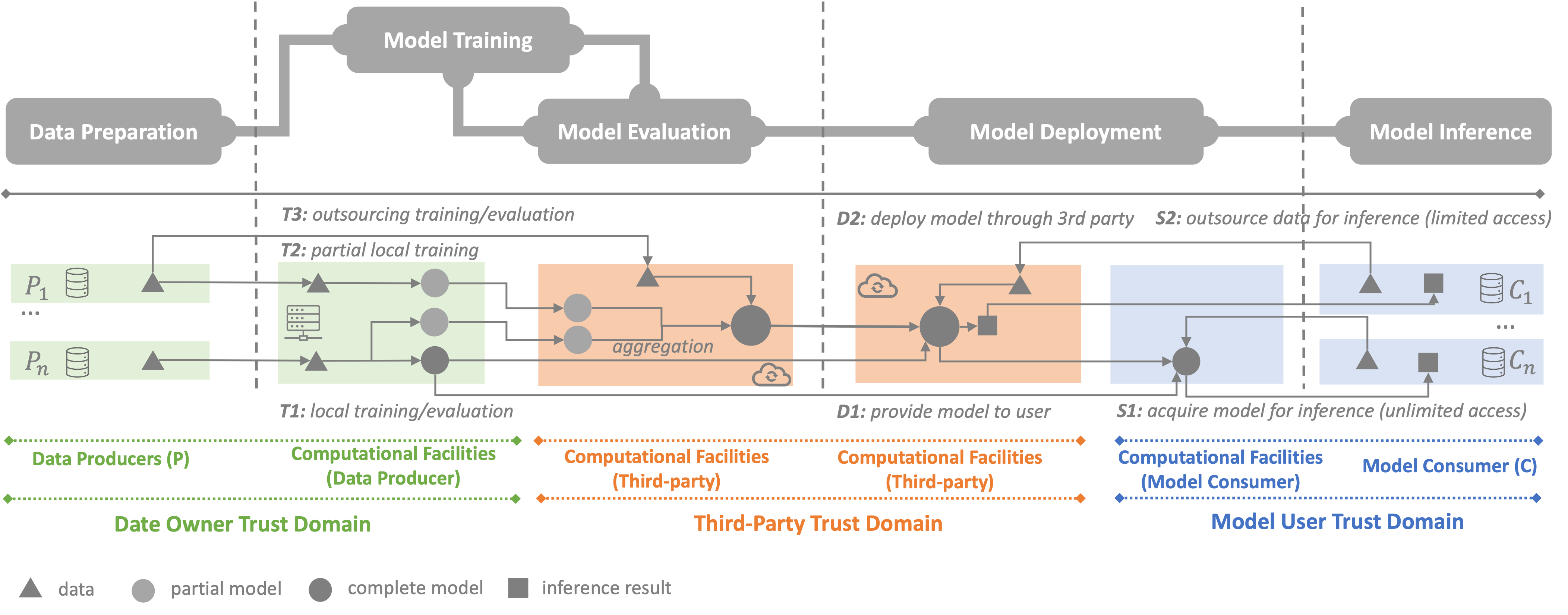}
    \caption{An illustration of machine learning pipeline (above part) and demonstration of corresponding processes showing different trust domains (bottom part) in various scenarios of ML applications.}
    \label{fig:chain}
\end{figure*}

% As depicted in \figurename~\ref{fig:chain}, we briefly overview the ML pipeline and the corresponding process chain and facilities, including self-owned and third-party computational resources (e.g., IaaS, PaaS, and MLaaS).
As seen in figurename~\ref{fig:chain}, we provide a high-level overview of the machine learning pipeline, its associated process chain, and relying facilities, which include the data owner's own devices and third-party-provided resources (e.g., IaaS, PaaS, and MLaaS).
Note that third-party facility-related ML pipeline is also a widespread adoption in recent ML-related applications.
The pipeline is divided into four stages: 
\textit{data preparation} stage where data is collected and preprocessed; 
\textit{model training and evaluation} stage where an ML algorithm is used to train an ML model and the trained model is evaluate; 
textit{model deployment} stage that involves steps to provision the model to the target user or deploy it through a third-party (e.g., as a service);
and \textit{model serving} stage where the model is used by the user to obtain the prediction/inference results.

From the perspective of the types of participants, the ML pipeline includes three entities/roles: the \textit{data producer (DP)}, the \textit{model consumer (MC)}, and the \textit{computational facility (CF)}; the CF may be owned by the data producer or model consumer themselves, or is employed by a trusted third-party.
The \textit{data producer} owns and provides the training data to train various ML models. 
% At the same time, the \textit{model consumer} also owns huge amounts of target data and expects to acquire various ML inference services such as classification of the target data with labels, prediction values based on the target data, and aggregation to several groups.
Simultaneously, the \textit{model consumer} possesses massive amounts of target data and expects acquiring various machine learning inference services such as labeling the target data, predicting values, and clustering the target data into groups.

For model training/evaluation or model deployment/serving stages, there exist two possible cases: the data producer (or the model consumer) (\romannumeral1) intends to use locally owned CFs, or (\romannumeral2) prefers to employ third-party CFs instead of local CFs.
As a result, different computation options exist for model training/evaluation and model deployment/serving.
As illustrated in \figurename~\ref{fig:chain}, in case (\romannumeral1), the \textit{data producer} can train a complete model locally (T1) or locally train a partial model that can be used to synthesize a global ML model in a collaborative or distributed manner (T2).
In case (\romannumeral2), the \textit{data producer} directly sends out its data to third-party entities that have computational facilities to employ their computational resources to train an ML model (T3).
Such third-party facilities may include the \textit{edge nodes} in an edge computing environment and/or IaaS servers in an cloud computing environment. 
Accordingly, the \textit{model consumer} may also acquire the trained model directly (D1) for model inference service with unlimited access (S1) if it has local computational facilities; otherwise, the \textit{model consumer} can also utilize a third-party facility where the trained model is deployed (D2) to acquire the prediction service (S2).

From the perspective of privacy-preserving phases in an ML pipeline, we can classify PPML research into two directions: privacy-preserving training phase including private data preparation and model generation and the privacy-preserving serving phase involving private model deployment and serving.
We discuss the affected phases of privacy-preserving approaches in Section~\ref{sec:phase} in detail.

It is also important to consider the trust domains to characterize the trust assumptions and related threat models. 
We divide the ML system into three domains: the data owner's local trust domain, the third-party CF trust domain, and the model user's trust domain. 
Based on those trust domains, we analyze various types of privacy guarantees provided by a PPML system.
We present more details in Section~\ref{sec:pg}.

From the perspective of underlying techniques, we decompose recently proposed PPML solutions to summarize and classify the critical privacy-preserving components to evaluate their potential utility impact.
Intuitively, the underlying privacy-preserving techniques such as differential privacy, conventional multi-party computation building on the garbled circuits and oblivious transfer, or customized secure protocols are widely employed in the PPML solutions.
Besides, various well-designed learning architectures have been broadly studied under specific trust domains and threat models. 
Furthermore, emerging advanced cryptosystems such as homomorphic encryption and functional encryption also show their promise for PPML with strong privacy guarantees.
More detailed taxonomy and analysis will be introduced in Section~\ref{sec:tu}.

\section{Privacy-Preserving Phases in PPML}
\label{sec:phase}

In this section we present existing PPML solutions considering the privacy-preserving phases in an ML pipeline.
\figurename~\ref{fig:chain} illustrates four phases in a typical ML pipeline: \textit{data preparation}, \textit{model training and evaluation}, \textit{model deployment} and \textit{model serving}.
Correspondingly, the existing PPML pipeline involves (\romannumeral1) \textit{privacy-preserving data preparation}, (\romannumeral2) \textit{privacy-preserving model training and evaluation}, (\romannumeral3) \textit{privacy-preserving model deployment}, (\romannumeral4) \textit{privacy-preserving inference}.
For simplicity, we analyze the PPML mainly by focusing on \textit{privacy-preserving model generation} covering phases (\romannumeral1-\romannumeral2) and \textit{privacy-preserving model serving} including phases (\romannumeral3-\romannumeral4).

\subsection{Privacy-Preserving Model Generation}
\label{sec:phase:creation}

Most privacy-preserving model generation solutions emphasize that the adopted privacy-preserving approaches should prevent the leakage of private information in the training data from leaking the trusted scope of the data sources.
In particular, the key privacy leakage issues to consider during model generation relate to \textit{data} and \textit{computation}; existing research address these through follow two key research questions:
\begin{itemize}
    \item[(\romannumeral1)] how to distill/filter the training data so as to minimize or completely remove any privacy sensitive information;
    \item[(\romannumeral2)] how to computationally process the training data in a privacy preserving manner.
\end{itemize}

\subsubsection{Privacy-Preserving Data Preparation}

From the perspective of \textbf{\textit{data}}, existing privacy-preserving approaches focus on the following directions: 
(\romannumeral1) adopting the traditional anonymization mechanisms such as $k$-anonymity\cite{sweeney2002k},  $l$-diversity\cite{machanavajjhala2007diversity}, and $t$-closeness \cite{li2007t} to remove the identifier information in the training data before using the data for training;
(\romannumeral2) representing the raw dataset using a surrogate dataset by grouping the anonymized data \cite{yang2019tradeoff} or abstracting the dataset by sketch techniques  \cite{li2019privacy,haddadpour2020fedsketch};
(\romannumeral3) employing differential privacy mechanisms \cite{dwork2008differential, dwork2010boosting, dwork2014algorithmic} to add privacy budget (noise) into the dataset to avoid private information leakage.

Specifically, in \cite{friedman2008providing}, Friedman et al. try to providing $k$-anonymity in the data mining algorithm, while the works in \cite{lefevre2006workload,kifer2006injecting} focus on the utility metric and provide a suite of anonymization algorithms to produce an anonymous view based on ML workloads.
Besides, recently, differential privacy mechanism has shown its promise in emerging DL models that rely on training on large datasets.
For example, Abadi et al. \cite{abadi2016deep} proposes a differentially private stochastic gradient descent approach to train a privacy-preserving DL model.
Among more recent work, McMahan et al. \cite{mcmahan2017learning} demonstrates that it is possible to train large recurrent language models with user-level differential privacy guarantees with only a negligible cost in predictive accuracy.
Recent parameter-transfer meta-learning (i.e., the applications including few-shot learning, federated learning, and reinforcement learning) often requires the task-owners to share model parameters that may result in privacy disclosure.
Proposals of privacy-preserving meta-learning such as 
those proposed by Xu et al. \cite{xu2019hybridalpha} and Geyer et al. \cite{geyer2017differentially}  
% presented in  \cite{geyer2017differentially,xu2019hybridalpha} 
address the problem of private information leakage in federated learning (FL) by proposing an algorithm to achieve client-sided (local) differential privacy.
In \cite{li2019differentially}, Li et al. formalize the notion of task-global differential privacy and proposes a differentially private algorithm for gradient-based parameter transfer that satisfies the privacy requirement as well as retains provable transfer learning guarantees in convex settings.

Thanks to recent successful work related to computing over encrypted data (i.e., practical computation over the encrypted data), training ML models on encrypted data is emerging as a promising approach to protecting privacy of training data.
Unlike the traditional anonymization mechanisms or differential privacy mechanisms that are still susceptible to the inference or de-anonymization attacks, such as demonstrated in \cite{wondracek2010practical,rahman2018membership,shokri2017membership,qian2016anonymizing}, wherein an adversary may have additional background knowledge, the encryption based approaches can provide a stronger privacy guarantees - called \textit{confidential-level privacy} in the rest of the paper. Hence these encryption based approaches are receiving more and more attention in recently \cite{truex2019hybrid,xu2019cryptonn,xu2019hybridalpha,gascon2016secure,hardy2017private,cheng2019secureboost,gilad2016cryptonets,chabanne2017privacy,nandakumar2019towards}, wherein the training data or the transferred model parameter is protected by cryptosystems while still allowing the subsequent computation outside of the trusted scope of the data sources.

\subsubsection{Privacy-Preserving Model Training}

From a \textit{\textbf{computation}} standpoint, existing privacy-preserving approaches are also correspondingly divided into two directions: 
(\romannumeral1) for the case that the training data is processed by employing conventional anonymization or differential privacy mechanisms, the computation involved during training is as is done in a vanilla model training; 
(\romannumeral2) for the case that the training data is protected via cryptosystems, due to the confidential-level privacy, the computation involved in privacy-preserving (i.e., crypto-based) training is a bit more complex than normal model training.
The demand of training computation over the ciphertext indicates that the direct use of traditional cryptosystems such as AES and DES is not applicable here, as those cryptosystems only secure data rather than operating the ciphertext.
That crypto-based training makes use of recently proposed advanced cryptographic schemes that primarily include homomorphic encryption \cite{gentry2009fully,van2010fully,brakerski2014leveled,martins2017survey,acar2018survey} and functional encryption\cite{boneh2011functional,goldwasser2014multi,abdalla2015simple,attrapadung2010functional,ananth2019optimal,abdalla2019single,abdalla2019decentralizing} schemes; these enable computation over the encrypted data.
In general, homomorphic encryption (HE) is a form of public encryption that enables computation over encrypted data without requiring the data to be decrypted.
In HE, the result of the computation remains encrypted and is the encrypted version of the result that would be obtained when the same computation is conducted on the original data.
Similarly, functional encryption (FE) is a generalization of public-key encryption in which the holder of a functional decryption key is able to learn a function of what the ciphertext is encrypting without learning the protected inputs themselves.

Note that. compared to the non-crypted approaches, training over encrypted data may involve an additional step - \textit{data conversion} or \textit{data encoding}.
The reason for this step is that the majority of those cryptosystems, such as multi-party functional encryption \cite{abdalla2015simple,abdalla2018multi} and BGV scheme \cite{yagisawa2015fully} (i.e., an implementation homomorphic encryption), are built on the integer group, whereas the training data preprocessed utilizing widely used methods such as feature encoding, discretization, normalization, or the model parameter exchanged is always in floating-point numbers.
It is worth noting that this is not a requirement for all crypto-based training methods.
For instance, an emerging implemented CKKS scheme \cite{cheon2017homomorphic} - an instance of homomorphic encryption - can support approximate arithmetic computation.

Typically, data conversion consists of two operations: encoding and decoding.
The encoding phase is typically used to transform floating-point values to integers, which enables the data to be encrypted and then used in cryptographic-based training.
On the contrary, the decoding step is used to recover the floating-point numbers from the trained model or crypto-based training result.
Without a doubt, the accuracy and efficiency of those rescaling processes are dependent on the conversion precision level.
In Section~\ref{sec:tu}, we will discuss the potential impact of data conversion in further detail.

\subsection{Privacy-Preserving Model Serving}
\label{sec:phase:serving}
There is no apparent distinction between privacy-preserving model deployment and model inference in the majority of PPML systems; so, we refer to the discussion in this section as privacy-preserving model serving.

Compared to privacy-preserving training, tackling privacy-preserving model serving challenges is relatively simpler from a computational standpoint, as demonstrated in Section~\ref{sec:overview:cmp}.
Except for the emerging machine learning models of complex deep neural networks, there are few studies devoted exclusively to privacy-preserving inference.
We observe that the majority of PPML solutions that make use of advanced cryptosystems (primarily homomorphic encryption and related schemes) are limited to privacy-preserving inference, as these crypto-based solutions are inefficient when applied to the complex and massive training computations of neural networks. We also note that these solutions primarily secure inference data samples, trained models, or both.

Additionally, another subfield of privacy-preserving model serving research focuses on privacy-preserving model querying or publication in cases when the trained model is deployed in a privacy-preserving manner, with the model consumer and model owner typically separated.
The primary question here is how to prevent an adversarial model user from inferring the private information from the original training data.
According to various model inference attack assumptions, an adversary has limited (or unlimited) access times to query the trained model.
Furthermore, the adversary possesses (or lacks) additional knowledge about the trained model specification, which is referred to as white-box (or black-box) attacks.
To address those inference attacks, a naive privacy-preserving strategy would be to restrict number of queries or to decrease the background information related to the disclosed model for a specific model user.
Beyond that, potential preventative approaches include the following: 
\begin{itemize}
    \item[(\romannumeral1)] \textit{private aggregation of teacher ensembles (PATE)} approaches \cite{papernot2016semi,papernot2018scalable,liu2020revisiting}, wherein the knowledge of an ensemble of ``teacher'' models is transferred to a ``student'' model, with intuitive privacy provided by training teachers on disjoint data, and strong privacy ensured by noisy aggregation of teachers' responses; 
    
    \item[(\romannumeral2)] \textit{model transformation} approach such as  MiniONN \cite{liu2017oblivious} and variant solutions as in \cite{rathee2021sirnn}, where an existing model is transformed into an oblivious neural network supporting privacy-preserving predictions with reasonable efficiency;
    
    \item[(\romannumeral3)] \textit{model compression} approach, especially applied in the emerging deep learning model with a large set of model parameters, where knowledge distillation methods \cite{hinton2015distilling,polino2018model} are adopted to compress the deep neural networks model.
    While knowledge distillation's primary objective is to minimize the size of the deep neural network model, it also provides extra privacy-preserving capabilities \cite{papernot2016distillation,wang2019private}.
    Intuitively, the distillation technique eliminates redundant information in the model and decreases the probability that the attacker may infer potential private information in the model via repetitive queries.
\end{itemize}

\subsection{Full Privacy-Preserving Pipeline}

The notion of a \textit{full privacy-preserving pipeline} is rarely mentioned in PPML proposals. 
Existing PPML solutions either enable \textit{privacy-preserving model generation} or are primarily concerned with \textit{privacy-preserving model serving}.
As demonstrated in Section~\ref{sec:overview:cmp}, the model inference computation tasks can be considered as a non-iterative and simplified form of model training procedures.
Thus, from the standpoint of the computation, privacy-preserving inference problems could be considered as a subset of privacy-preserving training problems.
% Existing PPML approaches that focus on privacy-preserving training based on secure computation techniques also indicate support for privacy-preserving inference, 
The majority of PPML systems that emphasize privacy-preserving training via secure computation techniques also indicate theoretical support for privacy-preserving inference,
such as in \cite{mohassel2017secureml,rouhani2018deepsecure,mirhoseini2016cryptoml,xu2019cryptonn,nandakumar2019towards,xu2021nn}; these, therefore, may be regarded as full privacy-preserving pipeline approaches.

We emphasize that PPML solutions that rely on privacy-preserving data preparation techniques such as anonymization, sketching, or differential privacy are typically incompatible with privacy-preserving inference.
The model inference aims at obtaining an accurate prediction for a single data point, those approximation or perturbation techniques are either inapplicable to the data required for prediction or lower the usefulness data for inference.
Thus, the data-oriented PPML proposals as introduced in Section~\ref{sec:phase:creation} are incompatible with the \textit{privacy-preserving inference} goal.

Another direction of a full privacy-preserving pipeline could be simply integrating privacy-preserving model generation approaches and those privacy-preserving model serving approaches; for instance, we can integrate a privacy-preserving model query or publication-based method for model deployment with a secure computation based privacy-preserving inference).
For instance, it is possible to produce a deep neural networks model with the most privacy-preserving training approaches. Then the trained model can be transformed into an oblivious neural network to support privacy-preserving predictions.

\section{Privacy Guarantee in PPML}
\label{sec:pg}

% Privacy is a jumbled idea.
% It's difficult to describe precisely what privacy entails.
Privacy, in general, is a broad term that encompasses the freedom of thought, control over one's body, seclusion in one's home, control over personal information, freedom from surveillance, protection of one's reputation, and protection from searches and interrogations \cite{solove2008understanding}.
Privacy, simply put, is a subjective estimate of the degree to which personal information can be revealed to untrustworthy domains/entities and how much personal information is publicly available.
It's difficult to define what privacy is and how to measure it, because privacy is a subjective concept with varying opinions or points of view.

Typically, in the digital realm, a widely accepted minimum level of privacy protection is that of the personal \textit{identity} \cite{sweeney2002k, dwork2008differential, dwork2009differential}.
Some common approaches to privacy include \textit{differential privacy}, mechanisms \cite{dwork2008differential, dwork2009differential} and \textit{k-anonymity} mechanism \cite{sweeney2002k} and its follow-up work such as \textit{l-diversity} \cite{machanavajjhala2007diversity}, \textit{t-disclosure} \cite{li2007t}. 
Specifically, differential privacy mechanisms try to conceal individuals from a dataset's output patterns, such as the output of specified functions queried from the dataset.
The basic principle of differential privacy is that if the effect of an arbitrary single change in database entries is modest enough, the query result cannot be used to infer much about any specific individual, and thus provides privacy protection \cite{dwork2008differential, dwork2009differential}.
The purpose of an anonymization procedure is to remove personally identifiable information directly from a dataset.
Given a person-specific field-structured dataset, k-anonymity and its variations are dedicated to creating and concealing identifiers and quasi-identifiers in such a way that the individuals who are the subjects of the data cannot be re-identified while the data remain useful \cite{sweeney2002k}.

Numerous privacy-related terminologies and concepts are used in PPML proposals to determine their privacy-preserving capabilities, resulting in the absence of a standardized definition of privacy in PPML.
We examine these commonly discussed privacy-related concepts in order to investigate PPML's privacy guarantees from two perspectives: \textit{object-oriented} and \textit{pipeline-oriented}.
The former focuses on evaluating the privacy protection of specific objects in PPML, namely, the trained model weights, exchanged gradients, and training or inference data samples.
The latter verifies the privacy assurance by evaluating the entire pipeline, as illustrated in \figurename~\ref{fig:chain}. 
Next, we expand on each perspective.

\subsection{Object-Oriented Privacy Guarantee}

% \subsubsection{Data Privacy}

The privacy claim of the majority of early PPML solutions is object-oriented, focusing on a single object such as a model or a data sample.
A series of PPML solutions directly protect the dataset, such as by empirically deleting \textit{identifiers} and \textit{quasi-identifiers} from the dataset using anonymization mechanisms \cite{sweeney2002k, machanavajjhala2007diversity,li2007t}, to meet the privacy goal.
To address the concerns raised by the above-mentioned privacy guarantee, a differential privacy (DP) mechanism \cite{dwork2008differential, dwork2009differential} has been developed and has been widely accepted across multiple domains as it provides a mathematically provable privacy guarantee.
Additionally, encryption is a more rigorous approach to data protection, requiring learning from the dark because the data is encrypted.
In the remainder of the paper, we generally refer to this type of privacy assurance as data-oriented privacy guarantee, or input privacy for short, as defined in  Definition~\ref{def:data_privacy}.

\begin{definition}[Data Oriented Privacy Guarantee]
\label{def:data_privacy}
\textit{A PPML solution asserts the data-oriented privacy promise, which states that an adversary cannot learn private information directly from input training/inference data samples or associate private information with a specific person's identification. }
\end{definition}

In short, data-oriented privacy-preserving approaches aim to prevent privacy leakage from the input dataset directly.
However, privacy is not free; one unintended consequence of input privacy is the sacrifice of data utility.
For instance, the anonymization mechanism needs to aggregate and remove proper feature values.
Simultaneously, certain values of quasi-identifier features are erased altogether or in part to fulfill $l$-diversity and $t$-disclosure definitions.
Additionally, a differential privacy technique requires the addition of a noise budget to the data sample.
Both methods have detrimental effects on the trained model's accuracy.
While encrypted data may ensure the dataset's confidentiality, it brings extra processing burden to the subsequent machine learning training.

% \subsubsection{Model Privacy}

% model privacy
Another group of PPML solutions focuses on delivering privacy-preserving models, implying that the trained model in the PPML system is the privacy-preserving model.
The privacy-preserving model's objective is to prevent privacy leaks in both the trained model and its use.
Examples of privacy information may include include information related to membership, property, attribute, etc., of a subject in a given data sample. 
As stated in Definition~\ref{def:model_privacy}, we refer to such a privacy assurance in the remainder of the paper as a \textit{model-oriented privacy guarantee} or \textit{model privacy} in short.

\begin{definition}[Model Oriented Privacy Guarantee]
\label{def:model_privacy}
\textit{A PPML solution is said to provide a model-oriented privacy guarantee if and only if an adversary cannot derive any private information from a given model by querying it a number of times.}
\end{definition}

Existing PPML approaches address model privacy guarantee using two approaches: 
(\romannumeral1) by incorporating differentially private training algorithms to perturb the trained model parameters; 
and (\romannumeral2) regulating the model access times and model access patterns to limit the adversary's ability to get private information.
For instance, Adabi et al. \cite{abadi2016deep} propose a differentially private stochastic gradient descent (DP-SGD) algorithm by adding a differential privacy budget (noise) into the clipped gradients to achieve a differentially private model. 
The private aggregation of teacher ensembles (PATE) framework \cite{papernot2016semi} creates an novel model deployment architecture in which a collection of ensemble models is trained as teacher models to offer model inference service for a student model.

\subsection{Pipeline-Oriented Privacy Guarantee}
\label{sec:pg:process}

Existing privacy measurement approaches such as differential privacy and $k$-anonymity are only applicable for certain entities such as data samples and trained models but cannot be directly adopted to assess the privacy guarantee of the entire PPML pipeline. 
There is a lack of formal or informal approaches for assessing the strength and scope of privacy protection provided by an ML pipeline.
We argue that assessing the privacy guarantee relies on defining  
(\romannumeral1) the \textit{boundary of data processing} 
and (\romannumeral2) the \textit{trust assumption on each processing domain} in the pipeline. 
For instance, suppose that a data owner employs a third-party computational facility (CF) to process its privacy-sensitive data; this creates a boundary in data processing workflow into two parts: data owner's local domain and CF's domain.  
If the data owner completely trusts CF, privacy concerns may not arise; otherwise, there is a need for a privacy guarantee with regards to data processing.

As illustrated in \figurename~\ref{fig:chain}, we establish the trust boundaries of  the processing pipeline in PPML as \textit{data producers}, \textit{local CF}, \textit{third-party CF}, and \textit{model consumers}.
From the perspective of a data owner (i.e., data producer), it may have varying levels of confidence in other domains.
For instance, the data producer may fully trust its local CF, or have semi-trust on the third-party CF, or may have no trust at all in the model consumer.
Based on such trust assumptions for each boundary, we present the taxonomy of privacy guarantees from the data owner's perspective as follows:

\begin{itemize}
    \item \textit{No Privacy Guarantee}: Here, the raw training data is shared with third-party CFs, regardless of their trustworthiness, and without using any privacy-preserving approaches. Each entity is able to acquire the original raw data to process or the ML model to consume according to its role in the ML pipeline.
    
    \item \textit{Global Model Privacy Guarantee}: Global privacy guarantee focuses on model serving phase.
    A data producer generates a trained model using its own CFs or enlists the assistance of a \textit{trustworthy} third-party entity with powerful CFs to assist in training the machine learning model using the raw data provided by the \textit{data producer}.
    The global privacy guarantee is designed to prevent the leakage of sensitive information during the model deployment and inference phases.
    In essence, the machine learning model is a statistical abstraction or pattern generated from the raw data, and hence any privacy leakage that occurs is considered as a statistical leakage. Typical types of privacy leakage include the disclosure of membership, class representatives, and properties. 
    For instance, a machine learning model for assisting in the diagnosis of HIV can be trained using current HIV patient healthcare records.
    A successful membership attack on the model enables an adversary to determine whether or not a specified target sample (i.e., patient record) was used in the training, hence disclosing whether or not a target person is an HIV patient.
    
    \item \textit{Vanilla Local Privacy Guarantee}: The basic local privacy guarantee ensures that the privacy-sensitive raw data is not directly shared with other \textit{honest} entities in the ML pipeline.
    The indirect-sharing approaches are as follows: (\romannumeral1) the raw data is pre-processed to remove privacy-sensitive information, or to obfuscate private information with noise before sending it out for model training; (\romannumeral2) the raw data is pre-trained in a local model, with the generated model update being revealed to other entities.
    
    \item \textit{Primary Local Privacy Guarantee}: The primary local privacy is built upon the vanilla local privacy guarantee.
    The primary distinction is in the trust assumption used to configure the remainder of the pipeline.
    Apart from the basic requirement of vanilla privacy guarantee, it also requires that the shared local model update should be protected from \textit{curious} third-party CF entities; here \textit{honest} CF entities may include the training server in the IaaS platform, and coordinating server in a distributed collaborative ML system.
    
    \item \textit{Enhanced Local Privacy Guarantee}: The enhanced local privacy is built upon the primary local privacy guarantee by changing the assumptions to inlcude the third-party CF entities that are totally \textit{untrusted}. 
    
    \item \textit{Full Privacy Guarantee}: The requirement of a full privacy guarantee includes both \textit{local privacy} and \textit{global privacy}.
    As the definitions as mentioned above, the \textit{global privacy} guarantee focuses on the ML model serving phase, while the \textit{local privacy} ensures the privacy guarantee in the model generation phase, the \textit{full privacy} ensures privacy protection at each step in the ML pipeline, as illustrated in \figurename~\ref{fig:chain}.  
\end{itemize}

In particular, the privacy leakage is more specific to the threat models considered that considers an adversary's behaviors and capabilities and assumes the worst-case scenario that a machine learning system can handle.
Simultaneously, the threat model also reflects users' confidence in the entire data processing pipeline and the trustworthiness of each entity.
As a result, as indicated previously, the privacy guarantees are highly correlated with the PPML system's specific threat model.

\section{Technical Utility in PPML}
\label{sec:tu}

We begin this section by classifying and discussing the PPML solutions in more detail by dissecting those solutions and examining how those approaches address the following questions:
\begin{itemize}
    \item How the privacy-sensitive data is released or published?
    \item How the privacy-sensitive data is used for model training?
    \item Does the architecture of the ML system prevent the disclosure of private-sensitive information?
\end{itemize}
As such, we summarize the privacy-preserving approaches into four categories: \textit{data publishing-based, data processing-based, architecture-based,} and \textit{hybrid approaches} that integrate two or three of those approaches.
Following that, we analyze the potential impact of these privacy-preserving techniques on normal ML solutions in terms of various utility costs, such as, \textit{computation utility}, \textit{communication utility}, \textit{model utility}, \textit{scalability utility}.

\subsection{Type I: Data Publishing Approaches}
\label{sec:tu:type1}

In general, the data publishing based privacy-preserving approaches in the PPML system fall into the following categories: 
approaches that \textit{totally eliminate} the identifiers and/or \textit{partially conceal} quasi-identifiers in the raw data; 
approaches that \textit{perturb} the statistical result of the raw data; 
approaches that \textit{completely transform} the raw data through the use of confusion and diffusion techniques.

\subsubsection{Elimination-based Approaches}

The traditional anonymization mechanisms are classified as elimination-based approach to prevent privacy leakage, in which techniques such as $k$-anonymity\cite{sweeney2002k},  $l$-diversity\cite{machanavajjhala2007diversity} and $t$-closeness \cite{li2007t} are applied to the raw privacy-sensitive data in order to eliminate private information.
Specifically, the $k$-anonymity mechanism aims to ensure that privacy sensitive information about an individual is indistinguishable from at least k-1 other individuals.
To accomplish this, $k$-anonymity defines the \textit{identifiers} and \textit{quasi-identifiers} for each data attribute, after which the identifiers are removed and the quasi-identifiers are partially obscured.
The $l$-diversity mechanism is based on $k$-anonymity by additionally maintaining the diversity of sensitive field, namely, \textit{equivalence class}.
An equivalence class has $l$-diversity if it has at least $l$ ``well-represented'' values for any privacy-sensitive attribute.
Essentially, as an extension of the $k$-anonymity mechanism, the $l$-diversity mechanism reduces the granularity of the data representation while maintaining the variety of sensitive fields through the use of techniques such as generalization and suppression, where any record can be mapped to at least $k-1$ other records in the dataset.
The $t$-closeness technique further refines the notion of $l$-diversity by imposing additional constraints on the value distribution on the \textit{equivalence class}; here, an equivalence class has $t$-closeness if the distance between the distribution of a sensitive attribute in this class and the distribution of the attribute in the entire dataset is less than a threshold $t$.

Examples of emerging elimination-based PPML solutions include approaches proposed by Yang et al. and Ong et al. in \cite{yang2019tradeoff,ong2020adaptive} that focus on secure or privacy-preserving federated gradient boosted trees model.
Yang et al. in \cite{yang2019tradeoff} employ a modified $k$-anonymity based data aggregation method to compute the gradient and hessian by projecting original data in each feature to avoid privacy leakage, instead of directly transmitting all exact data for each feature.
Additionally, Ong et al. in \cite{ong2020adaptive} propose an adaptive histogram-based federated gradient boosted trees by a data surrogate representation approach that is compatible with either the $k$-anonymity method or differential privacy mechanism.

\subsubsection{Perturbation-based Approaches}
We discuss two commonly used perturbation-based approaches: differential privacy mechanisms and sketching techniques.

\noindent\textit{\textbf{Differential Privacy}}: 
Typically, the perturbation-based privacy-preserving data publishing approaches primarily refer to \textit{$(\epsilon, \delta)$-differential privacy} technique \cite{dwork2008differential,dwork2010boosting,dwork2014algorithmic} and more recent \textit{$(\alpha, \epsilon)$- R\`{e}nyi differential privacy} \cite{mironov2017renyi} that is based on R\'{e}nyi divergence.
According to \cite{dwork2009differential,dwork2010boosting}, differential privacy is formally defined as follows:
a randomized mechanism $\mathcal{M}:\mathcal{D}\to\mathcal{R}$ with domain $\mathcal{D}$ and range $\mathcal{R}$ satisfies $(\epsilon, \delta)$\textit{-differential privacy} if for any two adjacent input $d,d^{'} \in \mathcal{D}$ and for any subset of outputs $S \subseteq \mathcal{R}$, it holds that
\begin{equation}
    \Pr[\mathcal{M}(d) \in S] \le e^{\epsilon}\cdot\Pr[\mathcal{M}(d^{'}) \in S] + \delta.
\end{equation}

The additive term $\delta$ allows for the possibility that plain $\epsilon$-differential privacy is broken with probability $\delta$ (which is preferably less than $1/|d|$).
Suppose that the R\'{e}nyi divergence of order $\alpha>1$ is defined as $D_{\alpha}(P||Q)$ over two probability distributions $P$ and $Q$. Similarly, the R\`{e}nyi differential privacy requires the following condition is hold:
\begin{equation}
    D_{\alpha}(\mathcal{M}(d)||\mathcal{M}(d^{'})) \le \epsilon.
\end{equation}
Usually, a paradigm of an approximating a deterministic function $f:\mathcal{D}\to\mathbb{R}$ with a differentially private mechanism is via \textit{additive noise} calibrated to function's \textit{sensitivity} $S_{f}$ that is defined as the maximum of the absolute distance $|f(d)-f(d^{'})|$.
The representative and common additive noise mechanisms for real-valued functions are Laplace mechanism ($\text{Lap}(\mu,b)$) and Gaussian mechanism ($\mathcal{N}(\mu,\sigma^2)$), as respectively defined as follows:
\begin{align}
\mathcal{M}_{\text{Gauss}}(d;f,\epsilon,\delta) 
&= f(d) + \mathcal{N}(\mu,\sigma^2) \\ 
% &= f(d) + \mathcal{N}(0,\frac{2\ln(1.25/\delta)}{\epsilon^{2}}\cdot S^{2}_{f})\\
\mathcal{M}_{\text{Lap}}(d;f,\epsilon) 
&= f(d) + \textit{Lap}(\mu,b)
% &= f(d) + \textit{Lap}(0,\frac{S^{2}_{f}}{\epsilon})
\end{align}

The typical usage of differential privacy in the PPML solutions falls into two directions:
(\romannumeral1) directly adopting the aforementioned additive noise mechanism on the raw dataset in the case of publishing data, as illustrated in \cite{chen2011publishing,jiang2013publishing}; or
(\romannumeral2) transforming the original training method into a differentially private training method so that the trained/published model provides $\epsilon$-differential privacy guarantee, as illustrated in \cite{abadi2016deep,geyer2017differentially,li2019differentially}.

\noindent\textit{\textbf{Sketching}}: \textit{Sketching} is an approximate and simple approach for data stream summarization, by empoying a probabilistic data structure that serves as an event frequency table, similar to counting Bloom filters.
Recent theoretical breakthroughs, such as in \cite{aggarwal2007privacy,melis2015efficient}, have demonstrated that with some adjustments, differential privacy is achievable through sketching techniques.
For instance, in \cite{balu2016differentially}, Balu and Furon focus on privacy-preserving collaborative filtering, a popular technique for the recommendation system, by utilizing sketching techniques to implicitly provide differential privacy guarantees by leveraging advantage of the inherent randomness of the data structure. 
Recently, Li et al. in \cite{li2019privacy} propose a novel sketch-based framework for distributed learning, where they compress the transmitted messages via sketches to achieve communication efficiency and provable privacy benefits simultaneously.

% a short summary here.
In short, the traditional anonymization mechanisms and perturbation-based approaches are designed to tackle general data publishing problems; however, those techniques are still not out-of-date in the domain of PPML.
The differential privacy mechanism, in particular, has been widely utilized in contemporary privacy-preserving deep learning and privacy-preserving federated learning systems, such as those proposed in \cite{abadi2016deep,mcmahan2017learning,geyer2017differentially,li2019differentially,truex2019hybrid,xu2019hybridalpha}.
Additionally, differential privacy is demonstrated not just as a privacy-preserving approach, but also in the generation of synthetic data \cite{dwork2008differential,jordon2019pate,triastcyn2018generating} and emerging generative adversarial networks (GAN) \cite{xie2018differentially,fan2020survey}.

\subsubsection{Confusion-based Approaches}
The confusion-based approach primarily refers to the \textit{cryptography} technique that \textit{confuses} the raw data in order to achieve a significantly greater privacy guarantee (i.e., confidential-level privacy) than typical anonymization mechanisms and perturbation-based approaches.
Existing cryptographic approaches for data publishing for ML training fall into following two directions:
(\romannumeral1) utilizing traditional symmetric encryption schemes such as AES in conjunction with the garbled-circuits and oblivious transfer to achieve general secure multi-party computation protocols \cite{mohassel2015fast,wang2017global,pinkas2009secure} and (authenticated) encryption in conjunction with the pairwise masking techniques \cite{bonawitz2017practical};
(\romannumeral2) utilizing advanced modern cryptosystems such as homomorphic encryption schemes \cite{gentry2009fully,van2010fully,brakerski2014leveled,martins2017survey,acar2018survey} and functional encryption schemes \cite{boneh2011functional,goldwasser2014multi,abdalla2015simple,attrapadung2010functional,ananth2019optimal,abdalla2019single,abdalla2019decentralizing} that contain the necessary algorithms to compute over the ciphertext, such that one party with the issued key is able to acquire the computation results.
The typical PPML system, such as those proposed in \cite{rouhani2018deepsecure,riazi2018chameleon}, could be classified as the first direction of the crypto-based data publishing approach, 
whereas more recent studies, such as those presented in \cite{truex2019hybrid,xu2019cryptonn,xu2019hybridalpha,gascon2016secure,hardy2017private,cheng2019secureboost,gilad2016cryptonets,chabanne2017privacy,nandakumar2019towards}, emphasize on the direction (\romannumeral2).

The confusion-based data publishing approaches (i.e., cryptographic-based systems), in particular, cannot work independently and are frequently coupled with subsequent secure process approaches, as the data receiver is meant to learn only the result of the data processing, rather than the raw data.
Although confusion-based approaches are promising candidates for data publishing, their introduction and discussion should focus on how to share the one-time symmetric encryption keys in direction (\romannumeral1) or how to process the encrypted data in direction (\romannumeral2).
The following section will go into extensive detail.

\subsection{Type II: Data Processing Approaches}
\label{sec:tu:type2}

The data processing approaches for training and inference are classified into two categories based on their respective data publication methodologies: \textit{ordinary computation} and \textit{secure computation}.
As with \textit{Type I} approaches discussed in Section~\ref{sec:tu:type1}, if the data is published using traditional anonymization mechanisms or perturbation-based approaches, 
in which personal identifiers in the data are eliminated and the statistical result is perturbed by adding differential privacy noise or constructing a probabilistic data structure, the consequent training computation is as normal as the training computation in vanilla machine learning systems.
Thus, the privacy-preserving data processing refers primarily to the secure computation that performed throughout the training and inference phases.

Andrew Yao \cite{yao1982protocols} initialized the secure computation problems and their accompanying solutions in 1982 using a garbled-circuits protocol for two-party computation challenges.
The major purpose of secure computation is to enable two or more parties to evaluate an arbitrary function of both their inputs without revealing anything except the function's output to either side.
According to the number of players enrolled, these secure computation approaches can be classed as basic secure two-party computation (2PC) and secure multi-party computation (MPC or SMC).
From the threat model's (a.k.a., security model's or security guarantee's) perspective, such secure computation protocols provide two distinct levels of security in response to varied adversary settings: \textit{semi-honest} (passive) security and \textit{malicious} (active) security.
We recommend the reader to \cite{cramer2015secure,hastings2019sok} for a detailed systematization of knowledge on secure multi-party computation solutions in general and their corresponding threat models.

We prefer to explore existing secure computation approaches in PPML in terms of the underlying technological principles in this section.
Generally, these secure computation approaches fall under the following categories:
\begin{itemize}
    \item[(\romannumeral1)] additive blindness with perturbation, DC-net, or (verifiable) secret sharing;
    \item[(\romannumeral2)] garbled-circuits technique with the oblivious transfer;
    \item[(\romannumeral3)] modern advanced cryptography schemes; 
    \item[(\romannumeral4)] mixed-protocols approaches; 
    \item[(\romannumeral5)] trusted execution environment technique with oblivious methods.
\end{itemize}
The underlying supported function could be generic or specific among various types of secure computation solutions.
The remainder of the paper will elaborate on each category of the solution.

\subsubsection{Additive Mask based Approaches}

A subcategory of secure computing approaches is additive blindness (or masking) techniques based on perturbation, DC-net, or secret sharing, in which private data are masked with randomized values that can be canceled out in the final computation output.

Additive perturbation is a straightforward type of additive masking.
For instance, in a generic additive perturbation based privacy-preserving summation \cite{sanil2004privacy} - a function-specific (i.e., the aggregation function) secure multi-party computation - the coordinator provides its input $x_0$ with adding a randomized perturbation $r$, and then each participant adds its input $x_i$ on the $x_0 + r$ and passes to the next participant. Finally, the coordinator receives the $\sum x_i + r$ and eliminate its randomized perturbation $r$ in order to obtain the aggregated result.
Another sort of additive masking is multiplicative perturbation, in which values are perturbed using random projection or random rotation techniques.

Additionally, in pairwise additive masking-based secure computing approaches, various secret sharing techniques such as $t$-of-$n$ secret sharing or multi-secret sharing are used.
For instance, assume a group of individuals desires to collaboratively compute the sum of their individual inputs, with the assistance of a semi-trusted entity (called coordinator).
For simplicity, we can assign the coordinator with a randomized nonce $s$ as the secret. Then each participant is issued with the secret sharing $s_i$ to add it into its input as a perturbation.
Finally, the coordinator can sum the $\sum x_i$ by removing the recovered secret $s$.
Recently proposed double-masking pairwise-based protocols \cite{bonawitz2017practical,kadhe2020fastsecagg,so2021turbo}  address participant failures by requiring pairs of participants to initially agree on pairwise masks via key exchange mechanisms.
Following that, each participant adds a self-mask and the sum of the pairwise masks of the other participants to its input.
In the recovery phase, the coordinator requests the alive participants with the sum of their (uncancelled) pairwise masks for the dropped users with added their ``self-mask'', and then subtracts those values from the previously masked input.
As a result, it can correctly compute the sum of the inputs of the undropped participants.

Additionally, anonymous communication could be a viable solution for pairwise blinding-based approaches.
For instance, dining cryptographer networks (DC-nets) \cite{chaum1988dining} or mix-nets \cite{chaum1981untraceable} are a sort of anonymous communication network in which only one person can send an anonymous message at a time.
In general, anonymous communication can be viewed as a restricted case of secure aggregation.
By utilizing DC-nets or mix-nets, the trusted coordinator can gather input from each party anonymously and then compute the function results, which provides a measure of privacy protection due to the coordinator's inability to determine the source of each function input.

In summary, the majority of additive blinding-based technique can be termed as lightweight approach in comparison to other forms of secure computing.
As illustrated in proposals \cite{agrawal2000privacy,kargupta2005random,kargupta2005random,bunn2007secure,doganay2008distributed,bonawitz2017practical}, those secure computing approaches are widely adopted in the traditional data mining area \cite{agrawal2000privacy,kargupta2005random} and lack attention in the recent privacy-preserving machine learning proposals.
Specifically, works such as demonstrated in \cite{bunn2007secure,doganay2008distributed} focus on the $k$-means clustering machine learning algorithms.
In \cite{bunn2007secure} Bunn and Ostrovsky propose two types of additive blinding methods, namely, a  \textit{division protocol} and a \textit{random value protocol} to perform two-party division and to sample uniformly at random from an unknown domain size.
Doganay et al. \cite{doganay2008distributed} utilize additive secret sharing as a cryptographic primitive to implement a secure multiparty computation protocol for privacy-preserving clustering.
Recent proposals, such as in \cite{bonawitz2017practical,kadhe2020fastsecagg,so2021turbo}, usually rely on a set of cryptographic primitives.
They employ a $t$-of-$n$ secret sharing scheme with additional DDH-based key agreement and authenticated encryption to construct a protocol for securely computing sums of vectors with low communication overhead, robustness to failures, and which requires only one server with limited trust.
To further improve computation and communication efficiency, Turbo-Aggregate \cite{so2021turbo} employs additive secret sharing and a multi-group circular strategy for secure aggregation tasks.
Simultaneously, FastSecAgg \cite{kadhe2020fastsecagg} proposes a novel multi-secret sharing scheme based on a finite-field version of the Fast Fourier Transform technique.

\subsubsection{Garbled Circuits based Approaches}
\label{sec:tu:type2:gc}

The garbled circuits and oblivious transfer techniques serve as the foundation for constructing another type of secure computing solutions.
To exemplify this, we utilize the two-party secure computation (2PC) protocol here.
The fundamental idea of 2PC is that one party (referred to as the garbled-circuit generator) creates a circuit computing function that includes a large number of garbled gates encrypted using typical symmetric encryption algorithms such as AES.
Then the other party (a.k.a, the garbled-circuit \textit{evaluator}) computes the output of the circuit obliviously, without learning any intermediate information.
Specifically, the function $f$ is transferred to a Boolean circuit comprised of vast amounts of garbled gates in various sorts (e.g., AND-gate, OR-gate, and XOR-gate).
Suppose that an AND-gate $g^{\text{AND}}$ is associated with two input wires $i$ and $j$, and one output wire $k$.
The generator first generates two cryptography keys for each input wire, denoted as $w^0_i,w^1_i,w^0_j,w^1_j$, where the superscript represents the the encoded input bits (e.g., $w^0_i$ encodes 0-bit input of wire $i$, while $w^1_i$ encodes 1-bit input of wire $i$).
For inputs data $b_i, b_j \in \{0,1\}$, the \textit{generator} computes the ciphertext as $\mathcal{E}^{\text{symmetric}}.\enc(w^{g^{\text{AND}}(b_i,b_j)}_k)$ with keys $w^{b_i}_i, w^{b_j}_j$.
\tablename~\ref{table:garbled_gate} presents the gate table in detail.
Then, the \textit{evaluator} is able to acquire its input wire associated keys $w^0_j,w^1_j$ with its input $b_j \in \{0,1\}$ without revealing that input to the \textit{generator} using the 1-of-2 oblivious transfer (OT) technique.
With the input associated key $w^{b_j}$ and the received permuted garbled table, the \textit{evaluator} is able to decrypt the corresponding ciphertext to acquire the output $w_k^{g^{\text{AND}}(b_i,b_j)}$ without learning the input of the \textit{generator}.
Finally, those different types of garbled gates can compose any functions used in the secure computation protocols.

\begin{table}
    \centering
    \caption{Illustrate of the garbled table for AND gate $g_{\text{AND}}$.}
    \label{table:garbled_gate}
    \scriptsize
    \begin{tabular}{cccccc}
        \toprule
            $b_i$ & $b_j$ & $g_{\text{AND}}$ & encrypted output & permutation & garbled output\\
        \midrule
            0 & 0 & 0 & $\enc_{w^{0}_i, w^{0}_j}(w^{0}_k)$ & $\Rightarrow$ & $\enc_{w^{0}_i, w^{0}_j}(w^{0}_k)$ \\
            1 & 0 & 0 & $\enc_{w^{1}_i, w^{0}_j}(w^{0}_k)$ & $\Rightarrow$ & $\enc_{w^{1}_i, w^{0}_j}(w^{1}_k)$ \\
            0 & 1 & 0 & $\enc_{w^{0}_i, w^{1}_j}(w^{0}_k)$ & $\Rightarrow$ & $\enc_{w^{0}_i, w^{1}_j}(w^{0}_k)$ \\
            1 & 1 & 1 & $\enc_{w^{1}_i, w^{1}_j}(w^{1}_k)$ & $\Rightarrow$ & $\enc_{w^{1}_i, w^{1}_j}(w^{0}_k)$  \\
        \bottomrule
    \end{tabular}
\end{table}

Even though garbled-circuits based 2PC and MPC problems is not an emerging topic and have been studied for over 40 years, the security community continues to work on them and attempts to improve their efficiency and practicality \cite{mohassel2015fast,ben2016optimizing,wang2017authenticated,wang2017global,katz2018optimizing}.
As a result of these efforts, the garbled-circuits based 2PC or MPC has been recently adopted to address the challenge of secure computation issues in popular machine learning algorithms, and more specifically, complex deep learning models \cite{gascon2016secure,mohassel2017secureml,liu2017oblivious,rouhani2018deepsecure,chandran2019ezpc}.
For example, Chameleon \cite{riazi2018chameleon} combines the best aspects of generic secure function evaluation protocols, where it employs additive secret sharing values to achieve linear operations and garbled-circuit protocols to implement nonlinear operations.
Similar to the Chameleon framework, $ABY^3$ \cite{mohassel2018aby3} proposes and implements a general framework for diverse machine learning algorithms in a three-server paradigm, based on the mixed 2PC presented by Demmler et al. \cite{demmler2015aby}, 
wherein data owners secretly share their data among three servers who train and evaluate models on the joint data using three-party computation.
Notably, several of those solutions, such as Chameleon and $ABY^3$, also make use of the homomorphic encryption technique, which will be discussed in later in this section.

DeepSecure \cite{rouhani2018deepsecure}, which is still based on Yao's garbled circuits, is a secure deep learning framework supporting various types of nerual networks that is built on automated design, efficient logic synthesis, and optimization methodologies.
Additionaly, Riazi et al. \cite{riazi2019xonn} propose another end-to-end framework based on Yao’s protocol that supports a paradigm shift in the conceptual and practical realization of privacy-preserving inference on deep neural networks.
DeepSecure's protocol is optimized for discretized neural networks with integer weights, whereas XONN is tuned for binary neural networks with boolean weights.
These quantized networks boost performance by eschewing expensive fixed-point multiplication in favor of integer or binary multiplication.
Recently, Agrawal et al. \cite{agrawal2019quotient} proposed QUOTIENT, a novel method for discretized DNN training along with a customized 2PC protocol.
EzPC \cite{chandran2019ezpc} is another sort of 2PC framework that generates efficient 2PC protocols from high-level, easy-to-write programs.
To achieve performance improvement, the proposed compiler of EzPC generates protocols0 combining both arithmetic and boolean circuits techniques.

Without relying on non-colluded two or three servers, those garbled-circuits-based solutions can provide provably security guarantee and demonstrate their promises in the training phase rather than merely the inference phase for deep neural networks, as exemplified in  \cite{rouhani2018deepsecure, agrawal2019quotient}.
Those systems, however, suffer from transmission overhead.
As illustrated in \tablename~\ref{table:garbled_gate}, to perform a simple computation on two input bits such as $b_i\wedge b_j$, it is mandatory to transmit a set of ciphertexts of fixed size and an additional oblivious transmission overhead for key delivery, where the size of each ciphertext and key depends on the secure parameter of the chosen symmetric encryption scheme.
Then, when complex computation functions such as those used in machine learning are considered, the size of transferred data explodes substantially.

\subsubsection{Advanced Cryptographic Approaches}
% \noindent\textbf{\textit{Modern Cryptosystem Based Approaches}}:
Another important direction of building secure multi-party computation is the modern cryptographic approaches, which mostly refer to advanced cryptosystems such as homomorphic encryption and functional encryption that enable computation over the ciphertext. 
Modern advanced cryptosystem-based secure computation can achieve a high level of privacy guarantee, as does the garbled-circuits-based approach, which makes use of the cryptosystem to provide confidentiality-level privacy.
Unlike garbled-circuits-based secure protocols, which are constrained by enormous amounts of transmitted data, modern advanced cryptosystem-based approaches require only encrypted data to be transferred, instead of the data-encoded garbled-circuits and corresponding keys via oblivious transfer technique.
Here we briefly introduce the \textit{homomorphic encryption} schemes \cite{gentry2009fully,van2010fully,brakerski2014leveled,martins2017survey,acar2018survey} and \textit{functional encryption} schemes \cite{boneh2011functional,goldwasser2014multi,abdalla2015simple,attrapadung2010functional,ananth2019optimal,abdalla2019single,abdalla2019decentralizing} that are primarily employed in existing PPML proposals \cite{xu2019cryptonn,hall2011secure,nikolaenko2013privacy,cock2015fast,gilad2016cryptonets,chabanne2017privacy,nandakumar2019towards,lou2019glyph,chabanne2017privacy}.

\noindent\textit{\textbf{Homomorphic Encryption (HE)}} is a public-key cryptosystem with the capacity of computing over ciphertexts without access to the private secret key. 
The result of the computation over the ciphertexts remains in the form of ciphertext. Simultaneously, the decrypted result corresponds to the outcome of operations performed on the original plaintext.
According to the capabilities of performing various kinds of operations, 
typical HE types include \textit{partially} homomorphic, \textit{somewhat} homomorphic, \textit{leveled fully} homomorphic, and \textit{fully} homomorphic encryption.
Unlike traditional public-key schemes, which consis of three primary algorithms: key generation (\textit{KGen}), encryption (\textit{Enc}), and decryption (\textit{Dec}), an HE scheme has an additional \textit{evaluation} (\textit{Eval}) algorithm for performing operations over ciphertext according to specified functions.
Formally, a HE scheme $\mathcal{E}_{\text{HE}}$ includes the preceding four algorithms such that 
% \begin{equation*}
% \label{eq:he}
%     \mathcal{E}_{\text{HE}}.Dec_{sk}(\mathcal{E}_{\text{HE}}.Eval_{pk}(f,\mathcal{E}_{\text{HE}}.Enc_{pk}(m_1), ..., \mathcal{E}_{\text{HE}}.Enc_{pk}(m_n))) = f(m_1, ..., m_n),
% \end{equation*}
\begin{align}
\label{eq:he}
    (\pk, \sk) &\leftarrow \mathcal{E}_{\text{HE}}.\kgen(1^{\lambda})\\
    \mathcal{C}_{\text{HE}} &\leftarrow \{\mathcal{E}_{\text{HE}}.\enc_{\pk}(m_i)\}_{i\in\{1,...,n\}} \\
    \mathcal{C}^{f}_{\text{HE}} &\leftarrow \mathcal{E}_{\text{HE}}.\eval_{\pk}(f,\mathcal{C}_{\text{HE}}) \\
    f(m_1, ..., m_n) &\leftarrow
    \mathcal{E}_{\text{HE}}.\dec_{\sk}(\mathcal{C}^{f}_{\text{HE}})
\end{align}
where $\{m_1, ..., m_n\}$ are the message to be protected, $\pk$ and $\sk$ are the key pairs generated by the key generation algorithm.

Typically, two representative approaches are available for achieving generic secure computation via HE techniques: \textit{preprocessing-model} approach and \textit{pure fully homomorphic encryption} (FHE) approach.
The former approach presupposes a trustworthy dealer, who does not need to know the function to be computed or the inputs and can be implemented through secure protocol with public-key infrastructure, but simply offers raw materials for the computation. 
Additionally, these operations can be performed as part of a preprocessing step using somewhat homomorphic encryption (SHE) techniques.
Following that, the online protocol evaluates a function securely by utilizing only low-cost information-theoretic primitives.
The \textit{pure FHE} approach, derived from the approach of FHE by Gentry \cite{gentry2009fully}, is more straightforward than the \textit{preprocessing model} approach.
In a pure FHE approach, all parties first encrypt their input using the FHE scheme and then use the homomorphic properties of the ciphertexts to evaluate the desired function on them.
Following that, these parties can conduct a distributed decryption operation on the final ciphertexts to obtain the results.

Here, instead of elaborating on all HE achievements, we briefly introduce commonly employed HE implementations shown in existing PPML systems.
We direct the reader to the article presented by Acar et al. \cite{acar2018survey} for the theory and implementation survey on HE schemes, as well as to an open consortium of HE standardization \cite{albrecht2018hestandard} to examine the availability of open-source libraries, for additional information.
The Paillier cryptosystem \cite{paillier1999public}, an additive homomorphic (partially homomorphic) encryption scheme, is one of the most extensively used HE implementations.
Given the message $m_i$ and $m_j$, the Paillier system $\mathcal{E}_{\text{HE}}^{\text{Paillier}}$ supports the additive homomorphic operation such that
\begin{equation}
    \mathcal{E}_{\text{HE}}.\enc(m_i) \circ \mathcal{E}_{\text{HE}}.\enc(m_j)=\mathcal{E}_{\text{HE}}.\enc(m_i+m_j)
\end{equation}
The HElib \cite{halevi2014algorithms} implemented various well-known fully homomorphic encryption schemes, including \cite{brakerski2014leveled,cheon2017homomorphic,smart2014fully,gentry2012homomorphic}, while also incorporating optimization techniques like as bootstrapping, smart-vercauteren, and approximate number.
SEAL \cite{sealcrypto2020} is another HE library that enables the computation of additions and multiplications on encrypted integers or real numbers. 
However, other operations, such as encrypted comparison, sorting, and regular expressions, are rarely possible to be evaluated on encrypted data using this library.
PALISADE \footnote{https://palisade-crypto.org/} is a more contemporary and general lattice cryptography library that presently supports efficient implementations of the following lattice cryptography capabilities including FHE schemes such as BGV \cite{brakerski2014leveled}, CKKS \cite{cheon2017homomorphic}, as well as multi-party extensions of FHE \cite{lopez2012fly}.

Several early proposals for privacy-preserving machine learning incorporate HE into regular machine learning models to prevent privacy leaking.
For instance, Hall et al. use homomorphic encryption to create a secure protocol for regression analysis in \cite{hall2011secure}.
Simultaneously, Nikolaenko et al. \cite{nikolaenko2013privacy} concentrate on privacy-preserving ridge regression on millions of records through the use of homomorphic encryption and garbled circuits.
Additionally, Cock et al. \cite{cock2015fast} present a computationally secure two-party protocol that is based on additive homomorphic encryption and eliminates the need for a trusted initializer.
Chialva and Dooms \cite{chialva2018conditionals} recently attempted to analyze the feasibility of homomorphic encryption being completely implemented in machine learning applications by addressing the comparison and selection/jump operations challenges.

HE is being used in conjunction with the success of emerging deep neural networks to achieve privacy-preserving deep learning.
For example, Gilad-Bachrach et al. \cite{gilad2016cryptonets} propose \textit{CryptoNets}, which aims to deploy neural networks over encrypted data by employing a leveled homomorphic encryption scheme to the training data.
\textit{CryptoNets} enable the addition and multiplication of encrypted data but require prior knowledge of the arithmetic circuit's complexity.
In contrast to the probable ineffectiveness of deeper neural networks in \textit{CryptoNets}, Chabanne et al. \cite{chabanne2017privacy} incorporate the batch normalization principle for the classification task using \textit{CryptoNets}' fundamental ideas.
Mishra et al. \cite{mishra2020delphi} recently proposed the Delphi framework for a cryptographic inference service for neural networks. They did so by constructing a hybrid cryptographic protocol that reduces communication and computation costs in comparison to previous work, as well as by developing a planner that generates neural network architecture configurations automatically.
As with Delphi, Lehmkuhl et al. \cite{lehmkuhl2021muse} propose the Muse framework to handle the challenge of fully malicious clients in secure inference scenarios rather than semi-honest clients.
Zheng et al. \cite{zheng2019helen} present \textit{Helen}, a framework for maliciously secure \textit{coopetitive} learning of a linear model without disclosing their data during the process of a distributed convex optimization technique called alternating direction method of multipliers (ADMM), in which a generic maliciously secure multi-party computation is based on the SPDZ protocol  \cite{damgaard2012multiparty}  derived from SHE schemes.
Following that Alexandru et al. \cite{alexandru2021encrypted} concentrate on encrypted distributed Lasso for sparse data predictive control using ADMM, which ensures the computational privacy of all data, including intermediate outcomes.
Additionally, Zheng et al. \cite{zheng2021cerebro} present Cerebro, an end-to-end learning platform that addresses the trade-off between privacy and transparency, as well as the trade-off between generality and performance.
Chen et al. \cite{chen2019efficient} present a multi-key homomorphic encryption scheme with packed ciphertexts and demonstrate how it can be used to securely evaluate a pre-trained convolutional neural network (CNN) model, in which a cloud server provides online prediction services to a data owner using a classifier provided by a model provider, while maintaining the privacy of both the data and the model.
Furthermore, Nandakumar et al. \cite{nandakumar2019towards} enable the secure training over deep neural networks using the open-source FHE toolbox HElib via a stochastic gradient descent training method.
Several more recent publications, such as \cite{crawford2018doing,lou2019glyph,hesamifard2019deep}, focus on the same problem but employ a variety of optimization techniques to improve model efficiency and accuracy.

\noindent\textit{\textbf{Functional Encryption (FE)}} is another type of public-key  cryptosystem that enables computation over the ciphertext.
Generally, a FE scheme $\mathcal{E}_{\text{FE}}$ consists of four algorithms: \textit{setup}, \textit{key generation}, \textit{encryption} and \textit{decryption} algorithms such that
% \begin{equation*}
%     \mathcal{E}_{\text{FE}}.Dec_{sk_{f}}(\mathcal{E}_{\text{FE}}.Enc_{pk}(m_1), ..., \mathcal{E}_{\text{FE}}.Enc_{pk}(m_n)) = f(m_1, ..., m_n),
% \end{equation*}
\begin{align}
    (\pk,\mathsf{msk}) &\leftarrow \mathsf{Setup}, \\
    (\sk_{f}) &\leftarrow \kgen(f, \mathsf{msk}), \\
    \mathcal{C}_{\text{FE}} &\leftarrow \{\mathcal{E}_{\text{FE}}.\enc_{\pk}(m_i)\}_{i\in\{1,...,n\}}, \\
    f(m_1, ..., m_n) &\leftarrow \mathcal{E}_{\text{FE}}.\dec_{\sk_{f}}(\mathcal{C}_{\text{FE}}),
\end{align}
where $\{m_1,...,m_n\}$ are the messages to be protected; the $\mathsf{Setup}$ algorithm creates a public key $\pk$ and a master secret key $\mathsf{msk}$, and $\kgen$ algorithm uses $\mathsf{msk}$ to generate a new functional private key $\sk_f$ associate with the functionality $f$.
% where the setup algorithm creates a public key $pk$ and a master secret key $msk$, and key generation algorithm uses $msk$ to generate a new functional private key $sk_f$ associate with the functionality $f$.
Typically, the algorithms of \textit{Setup} and \textit{KGen} usually are run by a trusted third-party authority.

Regarding implementations of FE, except for CiFEr\footnote{https://github.com/fentec-project/CiFEr}, PyFE\footnote{https://github.com/OpenMined/PyFE}, and FE-related PPML open-source projects: NN-EMD\footnote{https://github.com/iRxyzzz/nn-emd} and Reading-in-the-Dark\footnote{https://github.com/edufoursans/reading-in-the-dark}, 
there is a dearth of well-known implementation libraries comparing to HE-related libraries such as HElib and SEAL.
Existing construction of functional encryption schemes for general functionality, such as those recently proposed in \cite{goldwasser2014multi,boneh2015semantically,waters2015punctured,garg2016candidate,carmer20175gen, lewi20165gen}, place a premium on theoretical feasibility or presence of functionality.
Only a few recent proposals, for example in \cite{abdalla2015simple, abdalla2018multi, agrawal2011functional, bishop2015function}, emphasize the simplicity and applicability of FE, although the functionality is limited to the inner-products.

%[Discuss the similarity and difference - FE & HE]
As mentioned previously, the primary similarity between the \textit{FE} and \textit{HE} is that they both permit computation over the ciphertext.
The primary distinction between functional and homomorphic encryption, at a high level, is who can obtain the disclosed computation result.
Given an arbitrary function $f(\cdot)$, homomorphic encryption allows computing \textit{an encrypted result of} $f(x)$ from an encrypted $x$.
In contrast, functional encryption allows computing \textit{a plaintext result of} $f(x)$ from an encrypted $x$ \cite{alwen2013relationship}.
Intuitively, the function computation party in the HE scheme (i.e., the evaluation party) can only contribute its computation power to obtain the encrypted function result but cannot learn the function result unless it has the secret key.
In contrast, the function computation party in the FE scheme (i.e., usually, the decryption party) can obtain the function result with the issued functional private key.
Besides, except for most recently proposed decentralized FE schemes \cite{chotard2018decentralized,abdalla2019decentralizing,chotard2020dynamic}, the classic FE schemes are relied on a trusted third-party authority to provide key services, such as issuing a functional private key associated with specific functionalities.
We recommend the reader to \cite{xu2020revisiting} for a thorough analysis of the unique properties of new and promising functional encryption approaches for a variety of secure computation workloads.

Unlike the widely used HE-based approaches for the PPML secure computation tasks, recently proposed FE-based PPML solutions, such as those in \cite{xu2019hybridalpha, xu2019cryptonn, ryffel2019partially}, are beginning to demonstrate their efficiency and applicability.
Ryffel et al. \cite{ryffel2019partially} present a viable method for performing partially encrypted and privacy-preserving predictions using adversarial training and functional encryption.
Using the FE to construct the secure computing mechanism, Xu et al. \cite{xu2019cryptonn} initialize a CryptoNN framework that facilitates training a neural network model over encrypted data.
Additionally, Xu et al. focus on privacy-preserving federated learning (PPFL) in \cite{xu2019hybridalpha}, where they leverage the FE to design a secure aggregation technique that protects each participant's input in the PPFL.

\noindent\textit{\textbf{Impact of Data Encoding}}. 
Unlike the computation in anonymized or deferentially private PPML training, which is performed in the same way as non-PPML computation over floating-point numbers, the secure computation in crypto-based PPML training should be performed in the integer format, as those cryptosystems are constructed in the integer group.
As a result, there is a process for converting the data to integer format prior to doing the secure computation and then recovering the result in floating-point numbers.
As a result of this procedure, a problem arises during cryptographically secure computation: how to determine the encoding degree and the resulting influence of encoding precision.
As demonstrated in part in \cite{xu2019cryptonn,xu2019hybridalpha,zhang2020batchcrypt}, the encoding issue is a trade-off problem, in which increased encoding precision implies increased model accuracy. In contrast, a higher encoding precision typically results in much more secure computation time (i.e., more training time, especially in the large scale of data training.)

\subsubsection{Mixed-Protocol Approach}

The mixed-protocols solution that combines the aforementioned techniques is another direction to achieve efficient and practical secure multi-party computation. 
The underlying principle of these mixed-protocol approaches is to evaluate computing operations in terms of their most efficient representations.
The additions and multiplications with an efficient representation as an arithmetic circuit can use a homomorphic encryption approach. 
By contrast, the comparisons with an efficient representation as a boolean circuit will use Yao's garbled circuits technique.

Representative mixed-protocol solutions include TASTY \cite{henecka2010tasty}, ABY \cite{demmler2015aby}, ABY$^3$ \cite{mohassel2018aby3}, Chameleon \cite{riazi2018chameleon}, CrypTen \cite{knott2020crypten}, Falcon \cite{wagh2020falcon}, etc.
Notably, while a portion of the solutions listed above were presented in Section~\ref{sec:tu:type2:gc} as the garbled-circuits approach, we revisit those frameworks in a mixed fashion here.
Henecka et al. \cite{henecka2010tasty} present the \textit{TASTY} compiler, which is capable of generating protocols using HE, efficient garbled circuits, and their combinations.
Demmler et al. \cite{demmler2015aby} propose the \textit{ABY} framework, a mixed-protocol framework for efficiently combining secure computation approaches that are based on arithmetic sharing, boolean sharing, and garbled circuits.
\textit{ABY} pre-computes all cryptographic operations and then efficiently converts between secure computing approaches using pre-computed oblivious transfer extensions.
Following that, Mohassel and Rindal then enhance the ABY framework and offer ABY$^3$ for privacy-preserving machine learning in a three-party environment \cite{mohassel2018aby3}.
Riazi et al. \cite{riazi2018chameleon} propose the Chameleon framework to improve performance in terms of computation and communication between parties by overcoming two limitations: using a semi-honest third-party to preprocess arithmetic triples rather than the oblivious transfer used in ABY and handling signed fixed-point numbers.

Recently, in order to promote secure MPC adoption in the machine learning domain, Knott et al. \cite{knott2020crypten} proposed the \textit{CrypTen} framework, which exposes popular MPC primitives via abstractions that are commonly appeared in various machine learning frameworks, such as tensor computations, automatic differentiation, and modular neural networks.
Li et al. \cite{li2019privpy} present \textit{PrivPy}, an efficient framework for collaborative data mining with privacy protection, with the goal of providing an elegant end-to-end solution for data mining programming.
\textit{PrivPy}, in particular, provides more practical Python front-end interfaces, covering a broad range of functions frequently used in machine learning.
Simultaneously, the core compute engine is built on secret sharing, provides efficient arithmetics, and supports SPDZ \cite{damgaard2012multiparty}, and ABY$^3$ \cite{mohassel2018aby3}.
Notably, \textit{PrivPy} does not provide a theoretical breakthrough in cryptographic protocols; instead, it creates a practical solution that enables elegant machine learning programming on mixed MPC frameworks while making the appropriate trade-offs between efficiency and security.
As a follow-up to \textit{PrivPy}, Fan et al. \cite{fan2021ppca} focus on the privacy-preserving principal component analysis (PCA) via demonstrating an end-to-end optimization of a data mining algorithm to run on the mixed-protocol MPC framework.
To further improve the efficiency, CRYPTGPU \cite{tan2021cryptgpu} is proposed to accelerate the mixed-protocol MPC computation via GPU.
Specifically, CRYPTGPU introduces a new interface to losslessly embed cryptographic operations over secret-shared values into floating-point operations that highly-optimized CUDA kernels can process for linear algebra.

\subsubsection{Trusted Execution Environment Approach}
\label{sec:tu:type3}

The trusted execution environment (TEE) is a technique for creating an isolated environment that operates on a separate kernel and incorporates security features such as code authentication, runtime state integrity, and confidentiality for its code, data, and runtime states kept in permanent memory.
As a result, it is capable of providing a trusted environment in which users can run their programs in an untrusted server.
Notably, TEE intends to reduce the trusted scope of computation resources from the owner (e.g., cloud computing provider) to the maker of computation facilities (e.g., CPU manufacturer).
The TEE technique, in particular, is a method for achieving secure (or trust) computation for recently proposed work such as those in \cite{chamani2020mitigating,law2020secure}.

The TEE approach is reliant on the presence of a secure hardware enclave.
Examples of hardware enclaves include Intel SGX \cite{mckeen2013innovative} and AME Memory Encryption \cite{kaplan2016amd}.
Apart from trusted computing, another critical feature of TEE is remote attestation, which enables a remote client to properly verify that certain software has been loaded into an enclave securely.
Prior to that, a secure channel between the enclave and the client is bootstrapped, in which the enclave receives the client's public key and provides the signed attestation report.

A disadvantage of the TEE-based approach is the possibility of side-channel attacks that take advantage of information gleaned through the hardware implementation rather than flaws in the proposed algorithm itself.
Example of exploited information in the side-channel includes timing information, power consumption, electromagnetic leaks or even sound that can provide an additional source of information.
As a result, the TEE-based approach is frequently used in conjunction with oblivious techniques.
Typically, the accessible memory addresses must be concealed by making the enclave execution oblivious to the secret data, which requires either employing an oblivious data structure \cite{wang2014oblivious} within the enclave or operating the enclave atop an ORAM \cite{stefanov2013path}.
For example, Law et al. \cite{law2020secure} present a secure collaborative XGBoost system that enables multi-party training and inference of XGBoost models. Training takes place in the cloud, and individual clients' data is not revealed to the cloud environment or to other clients.
Additionally, they modify XGBoost's algorithms to be data-agnostic to avoid potential side-channel risk.
Similarly, Chamani and Papadopoulos \cite{chamani2020mitigating} present two modified versions of \textit{SecureBoost} \cite{cheng2019secureboost} that address the partial privacy leakage induced by sample partial ordering when the active party or all parties have access to the TEE.

Except for the TEE-based secure boost model examples discussed above, Cheng et al \cite{cheng2021separation}  recently propose \textit{Truda} that targets on a cross-silo FL system.
\textit{Truda} utilizes a decentralized and trustworthy aggregation architecture to alleviate information concentration around a single aggregator, in which all shared model updates in FL are disassembled at the parameter level and re-stitched to random partitions designated for multiple TEE-protected aggregators.
As a result, each aggregator only has a fragmentary and shuffled view of model updates and is oblivious to the model architecture.
Additionally, Zhang et al. \cite{zhang2021citadel} and Mo et al. \cite{mo2021ppfl} address the issue of scaling TEE-based distributed machine learning systems to large models or training datasets while remaining constrained by the capacity of the enclave page cache (EPC) or memory.
Specifically, Zhang et al. \cite{zhang2021citadel} propose a TEE-based method to divide ML training into training and aggregating parts, making it possible to spin up a distributed cluster to accommodate voluminous multi-sourced data.
Mo et al. \cite{mo2021ppfl} propose a PPFL solution, where both parties and aggregator own the TEE hardware for local training and global aggregation, respectively, by adopting greedy layer-wise training and aggregation to overcome the constraints posed by the limited TEE memory. 
As a result, it can provide comparable accuracy of complete model training but with the price of a tolerable delay.

\subsection{Type III: Architectural Approaches}

\begin{figure*}[t]
    \centering
    \includegraphics[origin=c,width=0.95\textwidth]{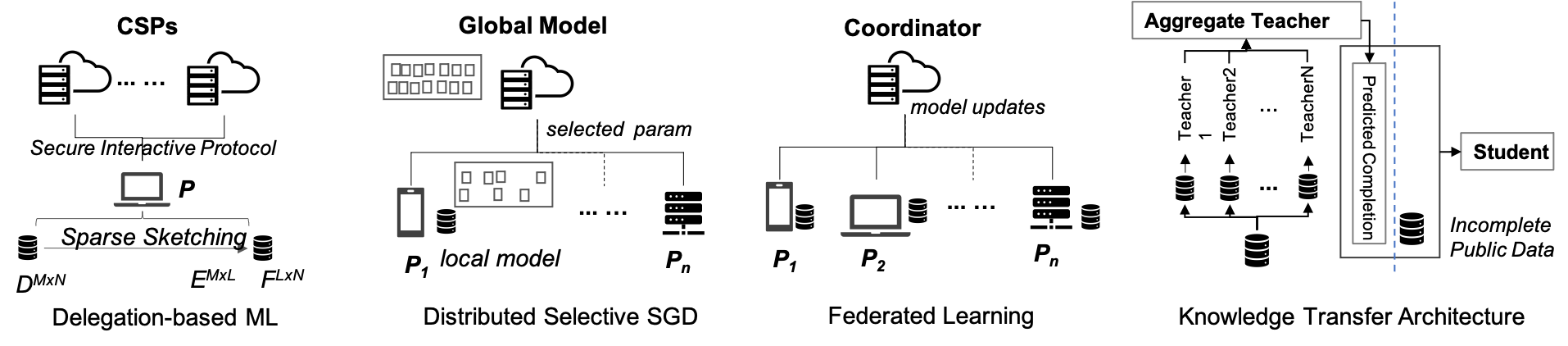}
    \caption{Representative architectures that have been incorporated into existing PPML solutions.}
    \label{fig:typical_arch}
\end{figure*}

Recently, a portion of promising privacy-preserving machine learning solutions has been achieved through intentionally or unintentionally designed architecture.
For example, PATE's architecture is primarily focused on knowledge transfer or knowledge distillation but also includes a guarantee of privacy for the disclosed model \cite{papernot2016semi}.
In short, there is no universally applicable design principle for privacy-preserving architecture-based approaches.
\figurename~\ref{fig:typical_arch} illustrates representative architectures used in existing PPML solutions to demonstrate how privacy can be ensured by designing a privacy-aware architecture with appropriate trust assumptions and threat model settings.
This section discusses representative architecture-related solutions for implementing PPML, rather than exhaustively listing all possible architecture-related privacy-preserving approaches.

\subsubsection{Delegation-based ML Architecture}
Delegation-based architecture is a classic architecture that gives the computation-limited parties the capability to create and use the ML models. 
Additionally, by incorporating additional secure techniques or making appropriate trust assumptions, the delegation-based architecture can provide privacy-preserving functionality for the machine learning system.
Mirhoseini et al. \cite{mirhoseini2016cryptoml}, for example, propose CryptoML, a practical framework that enables provably secure and efficient delegation for contemporary matrix-based machine learning systems, in which a delegating client with memory and computational resource constraints can assign storage and computation to the cloud via an interactive delegation protocol based on provably secure Shamir's secret sharing.
Similarly, Li et al. \cite{li2018privacy} propose a framework for privacy-preserving outsourced classification in a cloud computing environment, in which an evaluator can securely train a classification model using multiple encrypted data sources with distinct public keys.
Rather than outsourcing computation to a single server, Mohassel and Zhang \cite{mohassel2017secureml} present SecureML, an efficient two-server model in which data owners distribute their private data to two non-colluding servers that use 2PC to train various models on the joint data using stochastic gradient descent.

\subsubsection{Distributed Selective SGD Architecture}
Shokri and Shmatikov \cite{shokri2015privacy} propose a distributed selective SGD framework that enables multiple parties to jointly learn an accurate neural network model without sharing their input datasets. The system consists of several participants, each of whom has their own private training dataset, and one parameter server that is responsible for maintaining the most recent values of parameters available to participants.
More precisely, the approach presupposes two or more people training concurrently and independently.
For each round of local training, participants obtain the most recent values of the most-updated parameters and integrate them into local gradients using the selected partial parameters.
Following the local training, each participant has complete control over which gradients, how many gradients, and how often they share with the parameter server.

\subsubsection{Federated Learning (FL) Architecture}
The FL \cite{mcmahan2016communication,konevcny2016federated} is also a distributed machine learning framework with a similar architecture to the distributed selective SGD approach \cite{shokri2015privacy}, in which each participant maintains a private local dataset of its mobile facilities and a coordinator (a.k.a., an aggregator or a central server, as used in a few papers) trains a shared global model using the local model updates generated by those participants.
Due to the fact that training data does not leave the domain of each participant, FL can provide a primary level of privacy guarantee, as training data may be sensitive to sharing.

The FL design, in particular, can be viewed as a generic paradigm for the distributed selective SGD architecture outlined above.
The FL framework requires that the participant downloads all parameters in the global model, trains it on a local dataset, and then uploads the full local model (update) to the coordinator server if the participant has not dropped out during the current training epoch.
In comparison, the distributed selective SGD approach allows the participant more discretion over which partial parameters are included in the trained local or global models. 
The distributed selective SGD technique is, in essence, a sketching version of the FL.

Notably, FL has developed into a promising machine learning topic, attracting several studies on efficiency, scalability, security, and privacy. However, the growing systematization of knowledge article and survey have succinctly summarized FL, and as a result, we will not elaborate on FL here.
For specifics, we direct the reader to \cite{li2019federated,li2020federated,yang2019federated,yin2021comprehensive} .

\subsubsection{Knowledge Transfer Architecture}
The knowledge transfer-related architecture's major objective is to focus on knowledge distillation, model compression, and transfer learning.
However, a portion of emerging knowledge transfer systems may include a guarantee of privacy.

The architecture of private aggregation of teacher ensembles (PATE) \cite{papernot2016semi} and its variant \cite{papernot2018scalable,liu2020revisiting} are representative knowledge transfer-based PPML solutions.
In general, the knowledge of an ensemble of teacher models (i.e., the models that were initially trained) is transferred to a student model (i.e., the model that will be used), with intuitive privacy provided by training teachers on disjoint data and strong privacy guaranteed by noisy aggregation of teachers' responses.

The architecture of private aggregation of teacher ensembles (PATE) \cite{papernot2016semi} and its variant \cite{papernot2018scalable,liu2020revisiting} is a representative knowledge transfer based PPML solution.
In general, the knowledge of an ensemble of ``teacher'' models (i.e., initially trained models) is transferred to a ``student'' model (i.e., model that will be used), with intuitive privacy provided by training teachers on disjoint data and strong privacy guaranteed by noisy aggregation of teachers' answers.
In particular, PATE improves upon a specific, structured application of knowledge aggregation and transfer techniques.
PATE specifically strengthens the guarantee of privacy by limiting student training to a limited number of teacher votes and revealing just the topmost vote after carefully adding random noise.
Additionally, PATE restricts students' access to their teachers, allowing their exposure to teachers' knowledge to be quantified and bounded meaningfully utilizing knowledge transfer techniques such as generative adversarial networks (GANs).

In contrast to PATE's intuitive privacy, other widely used knowledge transfer techniques include \textit{model transformation} and \textit{model compression}.
MiniONN \cite{liu2017oblivious} and variants \cite{rathee2021sirnn}, for example, turn an existing model into an oblivious neural network capable of making privacy-preserving predictions.
Existing deep learning models in the natural language processing area with a large number of model parameters can be reduced to create lightweight deep learning models using knowledge distillation techniques \cite{hinton2015distilling,polino2018model}.
Apart from reducing the size of the deep learning model, as demonstrated and analyzed in \cite{papernot2016distillation,wang2019private}, it can also bring extra privacy-preserving functionalities.
Recently, it has been demonstrated that model compression and neural network pruning can be employed to achieve privacy-preserving functionality \cite{huang2020privacy,wang2021datalens}.
Huang et al. \cite{huang2020privacy}, for example, establish a link between neural network pruning and differential privacy.
Additionally, Wang et al. \cite{wang2021datalens} introduce the \textit{DataLens} framework, where they describe a scalable privacy-preserving training approach based on gradient compression and aggregation.
In comparison to PATE, which only allows ensemble teachers to vote on one-dimensional predictions, \textit{DataLens} combines top-k dimension compression with a related noise injection method to enable voting on high-dimensional gradient vectors while maintaining privacy.

\subsection{Type IV: Hybrid Approaches}
\label{sec:tu:type4}

Due to increased privacy protection requirements and recently demonstrated privacy threats such as membership inference attacks \cite{shokri2017membership,bernau2019assessing,jia2019memguard,li2020membership,nasr2018machine}, model inversion attacks \cite{fredrikson2015model,wu2016methodology,wang2015regression,he2019model}, and deep gradient leaking \cite{zhu2020deep,zhao2020idlg,geiping2020inverting}, implementing a single type of privacy-preserving technique outlined above is insufficient in some cases.
To achieve a higher level of privacy guarantee, an increasing number of well-built PPML systems incorporate more than one of the methodologies outlined above within an architecture that is appropriately tailored for the application scenario and threat model.

For example, the existing FL framework provides just a rudimentary guarantee of privacy, as each participant can save training data locally.
The global trained model, however, cannot withstand these membership inference and gradient inference attacks.
To solve this issue, as described in  \cite{abadi2016deep,geyer2017differentially,papernot2018scalable}, one type of hybrid method attempts to blend architecture-based approaches and differential privacy mechanisms.
Similarly, Liu et al. \cite{liu2019enhancing} propose exchanging sketched model updates rather than typical local model updates, as most FL systems do, because they recognize that sketching algorithms have the unique advantage of providing both privacy and performance benefits while retaining accuracy.
Additionally, to avoid the curious coordinator investigating the participants' input in FL and to improve the global model performance, another type of hybrid approach, as proposed in \cite{truex2019hybrid,xu2019hybridalpha,ma2021privacy}, integrates crypto-based secure aggregation approaches and differential privacy mechanisms into the FL framework to provide a stronger privacy guarantee, where secure aggregation approaches of \cite{truex2019hybrid,xu2019hybridalpha,ma2021privacy} are respectively based on partially additive homomorphic encryption, functional encryption, multi-key homomorphic encryption.
Apart from the hybrid study of differential privacy and cryptography in privacy-preserving federated learning systems, a new comprehensive study \cite{wagh2021dp} presents more examples of the conjunction of differential privacy mechanisms with cryptography schemes in generic domains.
We refer the reader to \cite{wagh2021dp} for a more extensive introduction divided into two categories: \textit{differential privacy for cryptography} and \textit{cryptography for differential privacy}.

Another type of hybrid method focuses on privacy guarantee during model training rather than  final  trained model via attempting to integrate the privacy-preserving architecture-based techniques and secure computation techniques.
For example, recent works, as those in \cite{hardy2017private,cheng2019secureboost}, employ secure multi-party computation and FL approaches to distribute training of machine learning models across many vertically partitioned datasets.
Likewise, these PPML systems \cite{mirhoseini2016cryptoml,li2018privacy,mohassel2017secureml} employ delegation-based architectures and secure computing methodologies.

In short, classic anonymization mechanisms and perturbation techniques (e.g., differential privacy) can protect the final trained model from the majority of model inference attacks.
Intuitively, the privacy-preserving approaches have obliterated or altered the private information contained in the training data samples; consequently, the final model cannot learn any privacy from the data.
Furthermore, in a distributed delegation-based machine learning scenario or in a FL paradigm, those systems cannot completely avoid the honest-but-curious central server investigating the users' input during the training phase.
Secure computation techniques such as garbled circuits or crypto-based 2PC and MPC can be used to protect the input of each participant.
They cannot, however, prevent the final trained model from leaking private information.

Recently, the FL paradigm has been combined with the trusted execution environment (TEE) approach to create privacy-preserving FL \cite{hashemi2021byzantine,zhang2021shufflefl,mo2021ppfl,cheng2021separation}.
For example, Hashemi et al. \cite{hashemi2021byzantine} propose constructing secure enclaves within the coordinator server of the FL paradigm by utilizing a TEE. Each client can then encrypt and transmit their gradients to verifiable enclaves.
Because the gradients are decrypted within the enclave, gradient-related privacy violations are avoided.
Additionally, Zhang et al. \cite{zhang2021shufflefl} present ShuffleFL, a method for defending against side-channel attacks that combines random group structure and intra-group gradient segment aggregation.
Mo et al. \cite{mo2021ppfl} address the issue of current TEEs having a limited memory space when employed in the FL by utilizing the greedy layer-wise training method to train each model's layer within the trusted area.
Cheng et al. \cite{cheng2021separation} propose \textit{Truda}, a cross-silo FL system that relies on a decentralized and trustworthy aggregation architecture to alleviate information concentration around a single aggregator. All shared model updates in FL are disassembled at the parameter level and re-stitched into random partitions designated for multiple TEE-protected aggregators.
As a consequence, each aggregator only sees a skewed and shuffled view of model updates and is unaware of the model architecture.

\subsection{Technical Approaches and Utility Cost}

\begin{table*}
    \centering
    \begin{threeparttable}
    \footnotesize
    \caption{Summary of primary technical path and utility cost of adopted privacy-preserving techniques in PPML solutions.}
    \label{table:tech_utility}
    \begin{tabular}{lll}
        \toprule
        Techniques & Primary Technical Design & Utility Cost\\
        \midrule
        $k$-anonymity ($l$-difersity, $t$-closeness) & anonymize private information  &  model utility\\
        differential privacy & perturb private information &  model utility \\
        sketching & sample from private information &  model utility\\
        compression & compress private information &  model utility \\
        homomorphic/functional encryption & diffuse/confuse private information &  computation utility\\
        pairwise additive mask (with SS, PKI) $\dagger$ & mask private information &  communication utility\\
        boolean garbled circuits (GC) & generic 2PC/MPC $\dagger$ &  communication utility \\
        boolean/arithmetic GC, SS, fully HE $\dagger$ & mixed 2PC/MPC $\dagger$ &  communication utility\\
        trusted execution environment & provide confidential computing &  scalability utility \\
        knowledge transfer architecture & prevent leakage from model &  scenario utility\\
        federated learning architecture & prevent data sharing & privacy strength utility \\
        \bottomrule
    \end{tabular}
    \begin{tablenotes}
        \footnotesize
        \item[$\dagger$] Abbreviation: PKI - public key infrastructure; SS - secret sharing; 2PC - secure two-party computation; MPC - secure multi-party computation. 
    \end{tablenotes}
    \end{threeparttable}
\end{table*}

In short, in comparison to present machine learning solutions, it is impossible to implement privacy-preserving machine learning without sacrificing utility, as discussed previously.
In this section, we summarize and discuss existing privacy-preserving approaches in terms of their \textit{primary technical design} and \textit{utility cost}.
The principal technical approach illustrates how these techniques fundamentally address privacy concerns, while the utility cost reflects the potential negative impact of employing these privacy-preserving techniques to achieve PPML solutions.

The fundamental technical design of existing privacy-preserving approaches is summarized in \tablename~\ref{table:tech_utility}, along with their possible utility cost.
The utility cost is composed of the following aspects: \textit{model utility}, \textit{computation utility}, \textit{communication utility}, \textit{scalability utility}, \textit{scalability utility}, \textit{scenario utility} and \textit{privacy strength utility}.
Specifically, typical anonymity strategies seek to eliminate identifiers or quasi-identifiers in order to prevent the leakage of private information, which may result in a reduction in model utility (i.e., model accuracy).
Similarly, privacy-preserving techniques that result in model utility costs include differential privacy mechanisms that inject noise into private information and approximation techniques such as sketching and compression techniques that are typically applied to intermediate model gradients.
The model cryptographic privacy-preserving approaches, such as homomorphic and functional encryption, provide confidential-level privacy and enable computation over ciphertext, resulting in a loss of computation utility.
Additionally, the generic 2PC or MPC offers secure computation through the use of boolean garbled circuits and oblivious transfer, which adds overhead to communication transmission.
Likewise, mixed-protocol systems optimize standard multi-party computation by incorporating additional techniques such as secret sharing, arithmetic corrupted circuits, and fully homomorphic encryption, but at the expense of communication and computation utility.
Emerging pairwise additive mask techniques concentrate exclusively on secure aggregation rather than generic secure computation, resulting in efficient computation due to the additive random mask protecting the private data. This system, however, requires multiple rounds of communication to agree on the random nonce and shared secret.
More recently, TEEs-based techniques provide a secure compute platform but suffer from limited hardware enclave size, limiting the scalability of machine learning solutions for large model sizes.
Additionally, the knowledge transfer architecture (e.g., PATE) is scenario-specific.
Due to the privacy leakage caused by intermediate model gradients, the federated learning paradigm is incapable of providing a solid guarantee of privacy.

\section{Challenges and Potential Directions}
\label{sec:cr}

Although the privacy-preserving topics were not fresh in the field of data mining \cite{agrawal2000privacy,lindell2000privacy}, privacy-preserving machine learning remains an active and ongoing research area due to
(\romannumeral1) the increasing adoption of traditional privacy protection mechanisms and recently proposed privacy-preserving techniques;
(\romannumeral2) rapidly evolving machine learning models such as emerging deep neural networks models; 
and (\romannumeral3) the emergence of stringent privacy-related policies and legislation.
Despite the fact that numerous privacy-preserving machine learning methods have been offered to meet compliance review and mitigate privacy concerns, a number of crucial challenges remain unexplored.
This section highlights open issues and challenges in PPML research prior to summarizing a few interesting research directions.

Notably, the open problems and challenges in vanilla machine learning systems also apply to the PPML domain, as a PPML system is constructed on top of a vanilla machine learning system.
However, this paper underscores the open problems and obstacles associated with privacy protection in the machine learning system.
As a result, we suggest readers to surveys or systemization of knowledge publications, such as those in \cite{kairouz2019advances, pouyanfar2018survey, li2019federated,yang2019federated}, for open topics and challenges in the vanilla machine learning research area, rather than replicating the content here.

\subsection{Open Problems and Challenges}
\label{sec:cr:challenges}

The evaluation of vanilla machine learning systems is primarily concerned on model performance and system efficiency, with model performance referring to model accuracy, robustness, and fairness, and system efficiency referring to training or inference time reduction.
Aside from that, existing PPML systems require additional work to ensure privacy.
Designing a well-designed PPML solution entails addressing the following open issues:

\begin{itemize}
    \item[(\romannumeral1)] In terms of privacy protection, how can a PPML solution be assured of adequate privacy protection in accordance with the trust assumption and threat model settings? Generally, the privacy guarantee should be as robust as possible from the data owners' standpoint.
    
    \item[(\romannumeral2)] In terms of model accuracy, how can we ensure that the trained model in the PPML approach is as accurate as the model trained in the contrasted vanilla machine learning system without using any privacy-preserving settings?
    
    \item[(\romannumeral3)] In terms of model robustness and fairness, how can we add privacy-preserving capabilities without impairing the model's robustness and fairness?
    
    \item[(\romannumeral4)] In terms of system performance, how can the PPML system communicate and compute as effectively as the vanilla machine learning system?
\end{itemize}

As the premise of a well-known adage - there is no such thing as a free lunch - implies, it is not free to power a vanilla machine learning system with privacy-protecting capabilities while keeping other system qualities such as model performance and efficiency.
In general, those open problems are mutually incompatible.
For example, in the vanilla deep neural networks model, increased model accuracy indicates the presence of numerous layers of neural networks, which requires additional training data and training epochs to achieve coverage.
As explained in Section~\ref{sec:tu}, using differential privacy as an example, implementing privacy-preserving techniques reduces model accuracy to a certain extent.
Specifically, Abadi et al. \cite{abadi2016deep} propose injecting the DP budget(noise) into the DP-SGD based deep learning model training; however, their evaluation results indicate that the model's accuracy cannot match that of the originally trained deep learning model.
Bagdasaryan et al. \cite{bagdasaryan2019differential} recently demonstrated a compatibility issue between privacy and fairness, showing and explaining that if the original model is unfair, the unfairness is exacerbated when the DP technique is used.

\begin{figure*}[t]
    \centering
    \includegraphics[origin=c,width=\textwidth]{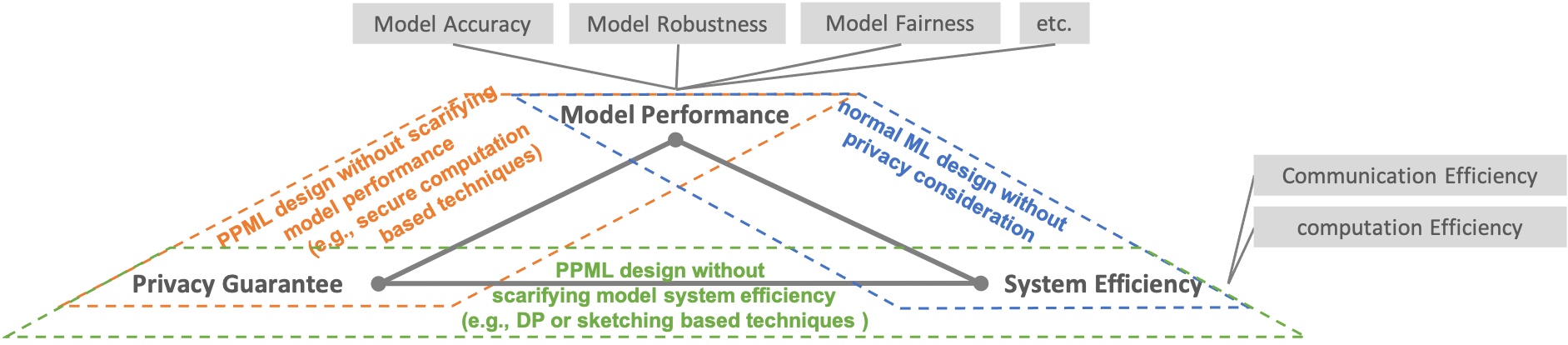}
    \caption{An illustration of the trade-offs that will be made when creating an optimal PPML solution among privacy assurance, model performance, and system efficiency.}
    \label{fig:trade_offs}
\end{figure*}

In summary, we believe that the primary challenge in developing an ideal PPML solution is resolving a trade-off dilemma between the solutions to those open problems, as illustrated in \figurename~\ref{fig:trade_offs}.
Existing PPML solutions trade either \textit{system efficiency} or \textit{model performance} for \textit{privacy guarantee}.
The efficiency of vanilla machine learning systems is concerned with how to improve training or inference efficiency, particularly for emerging deep neural network models with complicated network structures.
The majority of the machine learning community's efforts have been directed either increasing computation power, training in a distributed manner, or parallel training in order to address these efficiency issues.
Additionally, as indicated in \figurename~\ref{fig:trade_offs}, the efficiency issue with PPML systems highlights the following two concerns:
\begin{itemize}
    \item[(\romannumeral1)] PPML's communication efficiency emphasizes the importance of securely computing a function with fewer communication interactions and low transmission overhead;
    
    \item[(\romannumeral2)] PPML's computation efficiency emphasizes the need of securely computing a function with a low computational overhead or an acceptable computation time in the context of complicated machine learning training.
\end{itemize}

Besides that, as discussed and summarized above, the majority of existing PPML approaches focus on providing some degree of privacy-preserving features to a specific machine learning system; however, there is still a dearth of systematic definitions of privacy guarantees in terms of threat models or trust settings.
That is the reason that we present our initial summary on the privacy guarantee notations in Section~\ref{sec:pg}.
We expect that the presentation of privacy guarantees and the corresponding trust assumption and threat model settings will shed light on future PPML research.
Indeed, achieving broad agreement on a privacy guarantee definition for PPML systems remains a challenge.

\subsection{Research Directions}
\label{sec:cr:future}

This section summarizes our insights on the future directions of research in cross-domain of privacy-preserving techniques and machine learning.

\subsubsection{Systematic Definition, Measurement and Evaluation of Privacy}

As discussed in Sections~\ref{sec:phase} and~\ref{sec:tu}, various PPML solutions have demonstrated their efforts to provide machine learning systems with privacy guarantees and to include formal or informal privacy analysis to prove the privacy guarantees they claimed.
However, neither those PPML solutions nor a set of criteria could give a framework for evaluating or classifying the degree of privacy guarantee.

While we have provided a descriptive measurement and summary of the privacy guarantee in Section~\ref{sec:pg} in terms of the trust assumption and threat model settings for each entity in the machine learning pipeline, we believe it is insufficient for the PPML investigation.
There is still a need for a formal and systematic definition of the PPML system's privacy guarantee, as well as a commonly accepted framework or approach for evaluating the degree of privacy protection that a PPML system can provide.

\subsubsection{Attack and Defense Strategies}

In general, the machine learning system is vulnerable to three types of attacks:
(\romannumeral1) poisoning attacks that compromise the integrity of the training dataset collection; (\romannumeral2) inference attacks that infer private information from an individual participant's training data (or intermediate model update) or the trained (or aggregated) model; and (\romannumeral3) evasion/exploratory attacks that cause the trained model to produce incorrect (targeted/untargeted) classification outputs or collect evidence.

In terms of poisoning attack, several typical attacks include, for instance, clean-label data poisoning attack \cite{shafahi2018poison} where the adversary is assumed not to change the label of any training data and hence the poisoning of data samples has to be imperceptible, dirty-label data poisoning attack \cite{gu2017badnets} where the adversary is assumed to introduce a number of data sample with miss-classified and desired target label into the training set, and model poisoning \cite{bagdasaryan2018backdoor} where the adversary can poison local model updates before sending them to the server or insert hidden backdoors into the global model.

Regarding the inference attack, typical examples include membership inference attacks \cite{shokri2017membership,bernau2019assessing,jia2019memguard,li2020membership,nasr2018machine}, in which an attacker can infer whether a specific patient profile was used to train a classifier associated with a disease; model inversion attacks \cite{fredrikson2015model,wu2016methodology,wang2015regression,he2019model} that can use black-box access to prediction models in order to estimate aspects of someone’s genomics information;
deep leakage from gradients \cite{zhu2020deep,zhao2020idlg,geiping2020inverting} that obtains the private training data from the shared gradients in the ML cases of computer vision and natural language processing tasks.

PPML research aims to prevent the leakage of private information, therefore acting as a defender to such attacks.
As the adage goes, ``know yourself and your enemy, and you will never be defeated.'', understanding the fundamentals of those attacks will guide the research group to incorporate appropriate privacy-preserving strategies into the PPML solutions.
Additionally, it is worthwhile to investigate novel privacy-preserving approaches or to enhance existing methods in light of newly demonstrated inference attacks.

\subsubsection{Communication Efficiency}

As discussed in Section~\ref{sec:tu}, techniques such as boolean or arithmetic garbled-circuits, secret sharing, and oblivious transfer are widely used to construct a secure generic multi-party computation approach, which has been recently used in PPML solutions to provide privacy protections for the participant's input.
Based on the decomposition of the garbled circuits, it can be observed that each input should be encoded with associated keys and form different types of permuted garbled tables.
As a result, this approach places a significant burden on protocol communication in terms of training and inference; this is especially true when dealing with recent deep neural network models that involve complex network architecture and rely on massive amounts of training data, such as the most recent GPT-3 language model that was trained over 45TB of data and contains 175 billion parameters \cite{brown2020language}.
Additionally, the new approach to secure aggregation based on pairwise masking and secret sharing relies on multi-round peer-to-peer communication.

Two directions could be considered here to increase the communication efficiency of secure multi-party computation.
The first direction is to optimize the PPML solution's computation procedures.
For example, we may want to minimize the computational complexity of traditional machine learning systems or to improve the architecture of deep neural networks without impairing model performance.
The second option may involve optimizing garbled-circuits MPC protocols, for example, by offering a well-designed compiler that generates fewer garbled-circuits gates and thus reduces data transmission size.
It is worthwhile to optimize the secret-sharing strategy or communication topology when it comes to pairwise masking-based secure aggregation.

\subsubsection{Computation Efficiency}

Similarly to the optimization of garbled-circuits-based secure computation approaches, the optimization of emerging crypto-based secure computation approaches can also focus on optimizing the machine learning model to reduce the computing burden.

Another alternative is to develop an efficient cryptography technique that enables computation over ciphertext, such as homomorphic or functional encryption schemes.
Specifically, despite the cryptographic community's efforts to propose various types of homomorphic- or functional-encryption schemes with the goal of providing formal security proofs, increasing security degrees, or supporting more generic functionality, there is still a dearth of efforts to construct simple, practical, functionality-specific encryption schemes.
We recommend the reader to \cite{xu2020revisiting} for a comprehensive study of the challenges and directions for secure computation using emerging and promising functional encryption approaches.
Particularly in the context of edge computing, where IoT sensors or mobile devices have low computational capabilities, such a demand emerges in the crypto-based PPML system.

Furthermore, with the exception of homomorphic encryption systems based on approximate number arithmetic, such as CKKS \cite{cheon2017homomorphic}, the majority of secure computation-related cryptography schemes perform on the integer group.
Thus, determining the encoding precision is another problem to resolve, as noted above, because the majority of crypto-based PPML systems rely on the conversion of integer and floating-point numbers between the cryptosystem and the machine learning system.
In general, less precise encoding implies more efficient secure computation, which results in poorer model accuracy, and vice versa.

\subsubsection{Privacy Perturbation Budget and Model Utility}

As previously stated, it is impossible to incorporate a privacy perturbation budget into a machine learning system without impairing model utility.
A prominent example is the recently popular privacy-preserving technique, $(\epsilon, \delta)$-differential privacy mechanism, in which $\epsilon$ denotes the privacy budget.
As proved in existing proposals such as \cite{abadi2016deep,yu2019differentially}, the privacy budget $\epsilon$ negatively correlates to the model accuracy in the DP-based PPML solution.
To be more precise, if the PPML system chases a higher privacy budget (a.k.a., a tighter privacy guarantee), it will diminish model accuracy; however, a lower privacy budget implies a greater likelihood of privacy inference attacks succeeding.

As a result, two prospective avenues for research are as follows:
(\romannumeral1) How can we determine an appropriate privacy budget in light of the likelihood of inference attacks and model accuracy, or are there any universal or dynamic approaches for determining the privacy budget for various machine learning models?
(\romannumeral2) How to minimize the local privacy budget for each participant in a distributed training environment without compromising the final global model's privacy budget?

\subsubsection{New Deployment Approaches of Differential Privacy in PPML}

The emerging adoption of differential privacy mechanisms in PPML focuses on two directions: 
(\romannumeral1) \textit{centralized differential privacy (CDP)} deployment approach in which the DP noise is injected by a centralized trust node that has access to all private data, 
and (\romannumeral2) \textit{local differential privacy (LDP)} adoption where each individual perturbs private data or corresponding ML model before sending out in distributed ML scenarios.  
An example of CDP adoption is the differentially private stochastic gradient descent (DP-SGD) based deep neural networks training \cite{bagdasaryan2019differential}.
Examples of the LDP method being used in PPML include \cite{geyer2017differentially,xu2019hybridalpha,truex2019hybrid}.
LDP's enhanced privacy qualities come at the expense of model utility.
As a result, it is worthwhile to investigate the new deployment of DP approaches in PPML solutions in order to ensure adequate privacy guarantees while increasing model utility via cryptographic primitives, anonymous communication techniques, or a newly designed machine learning training architecture.

\subsubsection{Compatibility of Privacy, Fairness, and Robustness}

Recent proposals \cite{bagdasaryan2019differential,cummings2019compatibility} demonstrate that using DP-SGD in neural network training has a disparate effect on model accuracy; specifically, accuracy reduces significantly more for underrepresented classes and subgroups, implying that DP-SGD exacerbates the model fairness issue.
In particular, it is impossible to establish differential privacy and perfect fairness while keeping non-trivial accuracy, even when we have access to the entire distribution of the data.
Regarding numerous other related methods, such as sketching algorithms, there is still a dearth of research into the feasibility of achieving both privacy and fairness concurrently.
Additionally, we must demonstrate why the differential privacy and fairness are incompatible.
Understanding the cause for compatibility may help in mitigating the DP mechanism's impact on the model's fairness.

On the other side, the relevance of privacy and model robustness still lacks sufficient exploring and study. 
Few recent studies, such as those in \cite{naseri2020toward,jagielski2020auditing,ma2019data}, have explored the impact of the DP mechanism on poisoning attacks against deep neural networks and backdoor attacks against federated learning, namely, to what extent DP can be used to protect not only privacy but also robustness in ML.
In short, it seems like both LDP and CDP can provide a substantially more vigorous defense against backdoor and poisoning attacks in practice than in theory. 
There is a dearth of research into the utility of alternative privacy-preserving approaches and their impact on the robustness of generic models.

\subsubsection{Novel Architecture of PPML}

Existing architecture-based PPML solutions, such as the federated learning paradigm or the PATE framework, have demonstrated promising results in terms of privately training models over independent datasets while resolving data silo issues.
However, there are still challenges to be addressed in the FL systems, such as inefficient communication, systems, statistical heterogeneity, and potential privacy leakages.
The reader is referred to \cite{li2019federated, yang2019federated, kairouz2019advances, lyu2020threats, yin2021comprehensive} for a comprehensive review of the FL systems' remaining open problems.
Additionally, in addition to existing FL \cite{mcmahan2016communication,konevcny2016federated} and PATE \cite{papernot2016semi,papernot2018scalable} paradigms, it is worthwhile to investigate novel distributed machine learning architectures for privacy preservation.

\subsubsection{New Model Publishing Method for PPML}

Differential privacy (DP) is the most widely used mechanism for publishing a (locally) trained machine learning model in the context of the DP-SGD training method, model aggregation in the FL system, or the machine learning as a service (MLaaS) architecture.
As noted previously, the DP's limitations are visible in the trade-offs between privacy budget and model accuracy, as well as the recently proposed compatibility issue.
Recent proposals to incorporate classic sketching techniques such as the count-min sketch - a time-honored approach for measuring network traffic or approximation query processing - into the FL system have demonstrated their promise in terms of privacy protection \cite{balu2016differentially, li2019privacy}.
It is still worthwhile to investigate different sorts of sketching techniques, as well as their application breadth and restrictions.
Apart from the DP and sketching techniques discussed previously, are there any other approximate methods that might be considered candidates for the privacy-preserving model publishing method?

\subsubsection{Interpretability in PPML}

Researchers in the ML domain have recently started to put efforts into the interpretability of deep learning models since the deep neural networks model is still not explainable to its results.
The interpretability study allows us to know the deep learning model, and hence it is also helpful to give us insights into how private information is disclosed.
As a result, the interpretability study may help employ proper privacy-preserving approaches or design new privacy-preserving methods.
Additionally, the PPML system itself lacks explanation, such as the interpretability of the impact and the effects of applying the privacy-preserving approach.

\subsubsection{Benchmarking}

Benchmarking, in the context of PPML, is the practice of comparing tools in order to identify the most performant PPML solutions, which does not receive adequate investigation.
It is worthwhile to construct a privacy-preserving benchmarking framework similar to LEAF \cite{caldas2018leaf}, a framework for comparing popular learning tasks, methods, and datasets in federated environments.
To evaluate PPML's performance in terms of privacy protection, model accuracy, fairness, and robustness, the framework could incorporate synthetic private datasets, privacy-preserving methods, inference attack toolkits such as the Adversarial Robustness Toolbox (ART)\footnote{https://github.com/Trusted-AI/adversarial-robustness-toolbox}, and common machine learning models.

\section{Conclusion}
\label{sec:conclusion}
In this paper, we summarize and discuss existing privacy-preserving methodologies used in machine learning systems from a variety of perspectives, including the phases of the machine learning system and the underlying design principles, as well as provide the corresponding most recent PPML proposals. Additionally, we evaluate the privacy guarantee by defining several levels of privacy guarantee in the PPML systems.
Finally, we have outlined the challenges and open problems as well as pointing out future directions.
Solving those problems will require interdisciplinary effort from machine learning, distributed systems, and security-privacy communities.

% \vspace{.1in}
% \noindent\textbf{Acknowledgement}.[TBA]

\bibliographystyle{unsrt}

\bibliography{references}
\end{document}